\documentclass[10pt]{article}

\usepackage{amsmath}
\usepackage{placeins}
\usepackage{placeins}
\usepackage{amssymb}
\usepackage{amsthm}

\usepackage{booktabs}       

\usepackage{graphicx}

\usepackage{cite}
\usepackage{color} 
\usepackage[colorlinks,breaklinks]{hyperref}
\hypersetup{hypertexnames=true,
  linkcolor=blue,anchorcolor=black,citecolor=blue,urlcolor=blue
}

\DeclareGraphicsExtensions{.png,.pdf}
\graphicspath{{./}} 
\def\sgn{\textrm{sgn}}

\usepackage[labelfont=bf,labelsep=period,justification=raggedright]{caption}
\usepackage{subcaption}

\makeatletter
\renewcommand{\@biblabel}[1]{\quad#1.}
\makeatother

\newcommand{\giacomo}[1]{}

\date{}

\begin{document}

\begin{flushleft}
{\Large
\textbf{Hardware-efficient on-line learning through pipelined truncated-error backpropagation in binary-state networks}
}
\\
Hesham Mostafa$^1$, 
Bruno Pedroni$^2$, 
Sadique Sheik$^3$,
Gert Cauwenberghs$^{1,2,3}$
\\
$^{1}$Institute for Neural Computation,$^{2}$Department of Bioengineering, $^{3}$BioCircuits Institute \\
UC San Diego, California, USA \\
E-mail: \{hmmostafa,bpedroni,ssheik,gert\}@ucsd.edu
\end{flushleft}

\begin{abstract}
Artificial neural networks (ANNs) trained using backpropagation are powerful learning architectures that have achieved state-of-the-art performance in various benchmarks. Significant effort has been devoted to developing custom silicon devices to accelerate inference in ANNs. Accelerating the training phase, however, has attracted relatively little attention. In this paper, we describe a hardware-efficient on-line learning technique for feedforward multi-layer ANNs that is based on pipelined backpropagation. Learning is performed in parallel with inference in the forward pass, removing the need for an explicit backward pass and requiring no extra weight lookup. By using binary state variables in the feedforward network and ternary errors in truncated-error backpropagation, the need for any multiplications in the forward and backward passes is removed, and memory requirements for the pipelining are drastically reduced.  Further reduction in addition operations owing to the sparsity in the forward neural and backpropagating error signal paths contributes to highly efficient hardware implementation.  For proof-of-concept validation, we demonstrate on-line learning of MNIST handwritten digit classification on a Spartan 6 FPGA interfacing with an external 1Gb DDR2 DRAM, that shows small degradation in test error performance compared to an equivalently sized binary ANN trained off-line using standard back-propagation and exact errors. Our results highlight an attractive synergy between pipelined backpropagation and binary-state networks in substantially reducing computation and memory requirements, making pipelined on-line learning practical in deep networks.


\end{abstract}
\section{Introduction}
\label{sec:introduction}
The immense success of artificial neural networks (ANNs) is largely due to the use of efficient training methods that can successfully update the network weights in order to minimize the training cost function~\cite{LeCun_etal15}. Backpropagation~\cite{Rumelhart_etal86b}, or gradient descent in multi-layer networks, has become the training method of choice as it provides a conceptually clear approach to minimizing the cost function that works very well in practice. Training ANNs using backpropagation, however, is still a computationally demanding task as successful learning requires several presentations of the training data to allow the backpropagation algorithm to slowly adjust the network parameters. ANN accelerators developed for deployment in low-power systems therefore typically do not implement the lengthy and power-hungry training phase and only implement the computationally cheaper forward/inference pass~\cite{Himavathi_etal07,Cavigelli_etal15,Chen_etal16,Han_etal16,Ardakani_etal16,Aimar_etal17}. These ANN accelerators can thus only implement pre-trained networks with fixed parameters.  While this approach is appropriate for ANNs that process data from sources whose statistics are known beforehand and from which large amounts of training data have been gathered in the past in order to pre-train the network, it is inappropriate in situations where the device has to interact with unexpected or new sources of data and has to build its own classification or inference model on the fly.

Typical approaches for implementing large-scale ANN accelerators with learning capabilities~\cite{Gomperts_etal11,Ortega_etal16} are based on ANNs with smooth activation functions and high precision weights and neuron values. This necessitates the use of Multiply and Accumulate (MAC) operations as well as hardware implementation of activation functions such as the hyperbolic tangent and logistic sigmoid. This makes the hardware implementation of the accelerator costly in terms of logic resources and memory. We make use of recent developments that show that ANNs with binary neurons nearly match the performance of ANNs with smooth activation functions~\cite{Hubara_etal16}. In a binary ($-1,+1$) neural network, the implementation of the activation function reduces to a comparator and the forward pass involves no multiplications.  In an unsigned ($0,1$) binary neural network, the number of operations is further reduced owing to the sparse neural representation, with additions only for the non-zero activations.  However, the backward pass in which the error from the top layer is backpropagated to deeper layers still involves multiplications. To avoid these multiplications, we describe an approximation to the backpropagation algorithm that truncates the error signal to a ternary ($-1,0,+1$) variable. This yields a training algorithm that performs well in practice and that does not require any multiplications, beside affording further savings in addition operations owing to sparsity in the backpropagated error signal.

State of the art ANNs typically have millions of parameters.
Optimizing the movement of these parameters between memory and the computational elements is a key step to improve the power-efficiency and speed of ANN accelerators. This is especially true if the network parameters or weights are stored off-chip as off-chip memory traffic could easily become the bottleneck limiting the accelerator speed. A straightforward implementation of online backpropagation on custom hardware would typically need to look up each network weight twice, once during the forward pass and once during the backward pass. The backward pass lookup is needed so that the weight can be used to backpropagate the errors and for the updated weight then to be written back into memory. Pipelined backpropagation is a technique that can be used to eliminate the extra weight lookup in the backward pass ~\cite{Petrowski_etal93}. In pipelined backpropagation, the backward learning pass used to push the errors from the network output to the network input and to update the weights is performed in parallel with the forward inference pass. This is achieved by maintaining a history of the network state and using this history to update the weights based on delayed errors. The length of this history, however, grows with the network depth. Previously proposed hardware implementations of pipelined backpropagation~\cite{Girones_etal05,Bahoura_Park11,Savich_etal12} therefore incur large memory overheads as the network depth increases. We show that the network history can be compactly represented in binary-state networks (BSNs) which drastically reduces the memory overhead needed to implement pipelined backpropagation, making it a viable option when training deep networks.

In section~\ref{sec:background}, we provide some background on the backpropagation algorithm and the architecture of BSNs. In section~\ref{sec:hw_impl}, we describe our version of pipelined backpropagation, the learning algorithm, and the hardware architecture implementing approximate pipelined backpropagation in BSNs. We present experimental results demonstrating the learning architecture embedded on an FPGA platform in section~\ref{sec:fpga} and present our conclusions and directions for future work in section~\ref{sec:conclusions}.

\section{Materials and Methods}
\subsection{Background}
\label{sec:background}
\subsubsection{Feedforward Networks and Backpropagation}
\begin{figure}[h]
  \centering
  \includegraphics[width=0.6\textwidth]{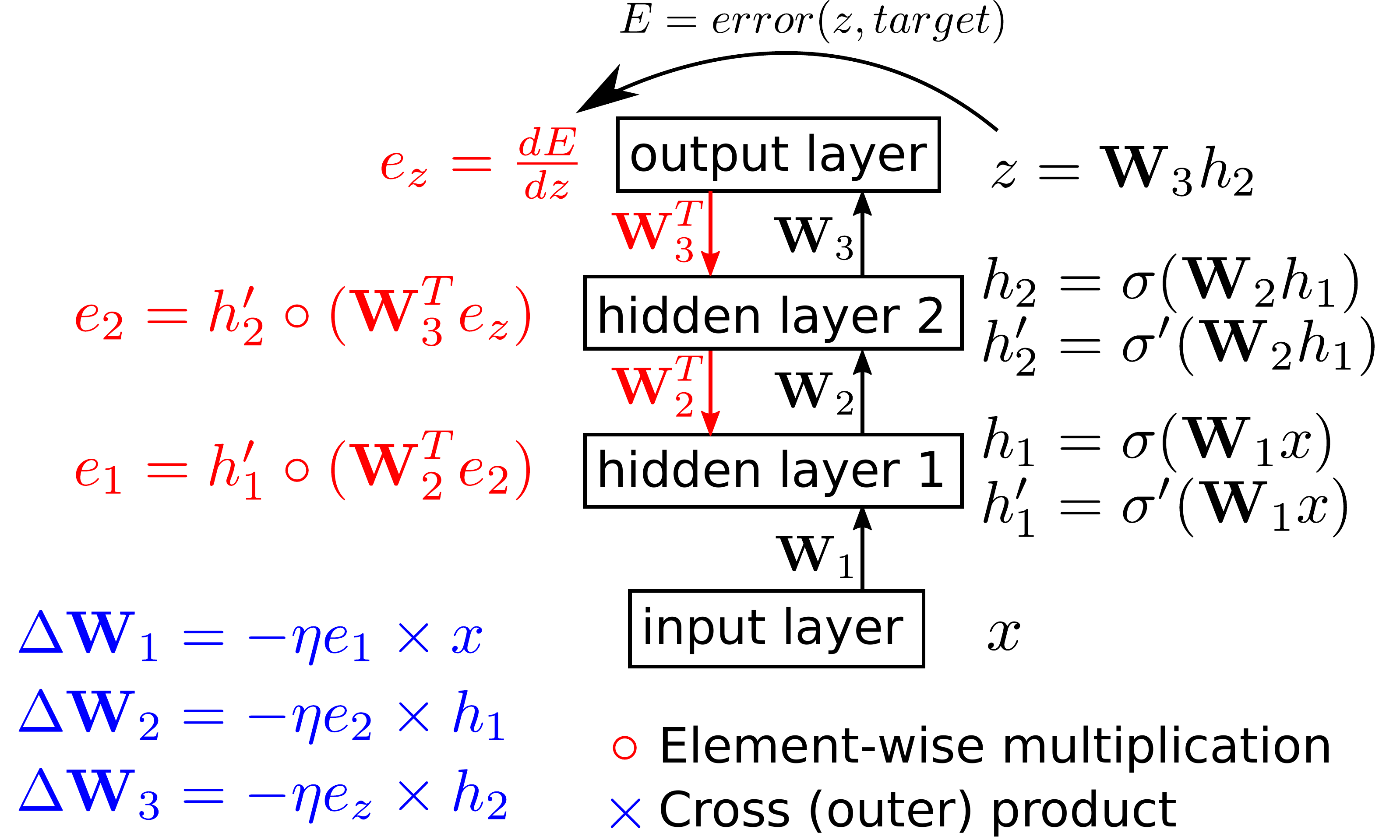} 
  \caption{The backpropagation algorithm in a feedforward network with two hidden layers. The biases in each layer are omitted for clarity. $\eta$ is the positive learning rate. $error(z,target)$ is the loss function and $target$ refers to the desired network output or the class of the presented input.}
\label{fig:backprop}
\end{figure}

Multi-layer fully-connected feedforward neural networks are typically arranged in layers. In a network with $L$ hidden layers, the activation vector, $h_i$, for hidden layer $i$ is given by
\begin{equation}
h_i = \sigma({\bf W}_ih_{i-1} + b_i) \quad i=1,..,L
\end{equation}
where ${\bf W}_i$ is the weight matrix connecting neurons in layer $i-1$ to neurons in layer $i$ and $b_i$ is the bias vector for layer $i$. $\sigma$ is a non-linear activation function that is applied element-wise on the argument vector. $x\equiv h_0$ is the input layer. The top layer in the network aggregates input from the last hidden layer to yield the network output. We use the vector $z$ to denote the top layer activity. The forward pass computation from network input to network output for a network with two hidden layer is depicted along the upward arrow in Fig.~\ref{fig:backprop}.

Network training involves updating the weights ${\bf W}_i$ and biases $b_i$ so as to minimize an error cost function $E$. The cost function quantifies the error in the network output and it is minimized when the network output approaches the desired output for the presented input. This cost function is almost always differentiable with respect to the network output. To minimize the cost function, gradient descent could thus be applied to move the network parameters along the negative direction of the cost function gradient with respect to the parameters. This yields the backpropagation algorithm which is illustrated in Fig.~\ref{fig:backprop} for the two-hidden layer network. To simplify the figure, the biases have been omitted. A layer bias vector can be implemented by having an extra neuron in the layer below that has no input and whose output is fixed at 1.

During the backward pass depicted in Fig.~\ref{fig:backprop}, the error derivative at the top layer, $\frac{dE}{dz}$, is pushed down through the network. The backward pass needs access to two sets of vectors: the activation vectors ($x$, $h_1$, and $h_2$ in Fig.~\ref{fig:backprop}) and the vectors containing the derivative of the activation function in the hidden layers at the operating point of the network during the forward pass ($\sigma'({\bf W}_1x)$ and $\sigma'({\bf W}_2h_1)$ in Fig.~\ref{fig:backprop}). The first set of vectors is needed to update the weights through a cross product with the error vectors and the second set of vectors (the derivative vectors) are needed to push the error from the top layer down through the layer stack.

\subsubsection{Binary-State Networks}
One of the surprising properties of neural networks is the resiliency of the backpropagation procedure to various deviations from the ideal algorithm depicted in Fig.~\ref{fig:backprop}. One hardware-oriented modification of the ideal backpropagation algorithm aims at training networks with binary activation functions~\cite{Hubara_etal16}. The neuron's binary output could be either $-1/1$ or $0/1$. The latter choice ($0/1$), which is equivalent to the bipolar binary representation ($-1/1$) through a linear transformation $h \leftarrow 2\, h - 1$, is preferable in application settings that benefit from greater sparsity in neural activity.  We investigate the performance of both formats and the implementation in Sec.~\ref{sec:hw_impl} supports both formats. The use of binary neurons considerably simplifies the forward pass through the hidden layers as the multiplications are replaced by sign change circuits ($-1/1$ format) or AND gates ($0/1$ format) . The neuron's activation function can be either $\sigma_{-1/1}$ or $\sigma_{0/1}$, which are given by:
\begin{equation}
\label{eq:binary_act_bipolar}
\sigma_{-1/1}(x) = \begin{cases}
   1 & \text{if \quad $x\geq 0$}  \\
   -1       & \text{otherwise.} 
  \end{cases}
\end{equation}
\begin{equation}
\label{eq:binary_act_unipolar}
\sigma_{0/1}(x) = \begin{cases}
   1 & \text{if \quad $x\geq 0$}  \\
   0       & \text{otherwise.} 
  \end{cases}
\end{equation}

The derivatives of $\sigma_{-1/1}$ and $\sigma_{0/1}$, however, are zero almost everywhere which would stop errors from backpropagating. A virtual derivative which was found to work well in practice is the saturating straight-through estimator~\cite{Hubara_etal16,Bengio_etal13}:
\begin{equation}
\label{eq:binary_der}
\sigma_{-1/1}'(x) = \sigma_{0/1}'(x) = \begin{cases}
   1 & \text{if \quad $-1\leq x \leq1$}  \\
   0       & \text{otherwise.} 
  \end{cases}
\end{equation}
If the input layer activity is also binarized, then the entire forward pass from input layer to output layer is free from multiplications. The backward pass, however, still involves multiplications when pushing the error down by one layer (the ${\bf W}_3^Te_z$ and ${\bf W}_2^Te_2$ operations in Fig.~\ref{fig:backprop}). In section~\ref{sec:hw_impl}, we describe how these multiplications can be avoided. 

A closely related development for reducing the computational and memory requirements of ANNs aims at training ANNs with low-precision weights~\cite{Stromatias_etal15b,Courbariaux_etal15,Rastegari_etal16,Zhu_etal16}. During training, a high-precision version of the weights is typically stored and updated based on the errors calculated in the network. During the forward and backward passes, a low-precision quantized version of the weights is used. After training, the network operates using the low-precision weights. Since high-precision weights are still maintained during training, this approach can not reduce the ANN memory requirements during training. However, the use of quantized weights reduces the complexity of the logic used to compute the forward and backward passes.

\subsection{Learning Algorithm and Hardware Architecture}
\label{sec:hw_impl}

\subsubsection{Learning Algorithm}
Our goal is to use binary fully connected feedforward networks, also known as multi-layer perceptrons, to solve classification tasks. The top layer has as many neurons as the number of classification classes. We impose a cost function on the activity of the top layer neurons that is minimized when the activity of the top layer neuron corresponding to the correct input class is the highest among the top layer neurons. Note that the top layer neurons are not binary. Two of the most popular cost functions used in classification settings are the cross entropy loss and the square hinge loss. Cross entropy loss, however, involves exponentials and logarithms while the square hinge loss requires multipliers to implement the squaring operations. Therefore we use a simpler loss function, the hinge loss. Let $z$ be the vector of top layer activity and $z[i]$ be the $i^{th}$ element of $z$. If $p$ is the index of the correct class and $C$ is the number of classes (also the length of $z$), then the hinge loss is given by:
\begin{equation}
  E_{hl} = \sum\limits_{\substack{i=1\\i \neq p}}^C \max(0,z[i] + H - z[p])
  \label{eq:hinge}
\end{equation}
where $H \ge 0$ is the hinge hyper-parameter. The gradient of $E_{hl}$ with respect to $z$ is thus given by:
\begin{equation}
  \frac{dE_{hl}}{dz[i]} =
    \begin{cases}
       \vartheta(z[i] + H - z[p]) & i \ne p \\
       -\sum\limits_{\substack{j=1\\j \neq p}}^C \vartheta(z[j] + H - z[p]) & i = p
    \end{cases}
    \label{eq:output_update}
\end{equation}
with the Heaviside operator $\vartheta(\cdot)$ defined as:
\begin{equation}
\vartheta(x) = \begin{cases}
   1 & \text{if $x > 0$ } \\
   0 & \text{otherwise.}    
  \end{cases}
\end{equation}
The gradient $e_z = \frac{dE_{hl}}{dz}$ can thus be efficiently computed using comparators and adders. Note that all elements of $e_z$ are in the range $[-(C-1),1]$.

As shown in the example in Fig.~\ref{fig:backprop}, the backward pass in a binary-state network (BSN) involves multiplications in order to compute ${\bf W}_3^Te_z$ and ${\bf W}_2^Te_2$. Since at most one element in $e_z$ can have a value other than zero or one, and the absolute value of this element is at most $(C-1)$, we can use repeated additions across at most $(C-1)$ cycles to calculate the product of this element with a weight. Computing the second term, however, is more challenging as $e_2$ can take a broad range of values. Thus, we modify the backpropagation scheme illustrated in Fig.~\ref{fig:pipelined_backprop} so that all errors below the top layer are truncated to -1, 0, or 1. In our running example in Fig.~\ref{fig:backprop}, $e_2$ and $e_1$ are modified so that
\begin{equation}
\label{eq:approx_error}
\begin{array}{c}
e_2 = \sgn(h_2' \circ {\bf W}_3^Te_z) \\
e_1 = \sgn(h_1' \circ {\bf W}_2^Te_2)
\end{array}
\end{equation}
where the signum operator $\sgn(\cdot)$ is defined as:
\begin{equation}
\sgn(x) = \begin{cases}
   1 & \text{if $x > 0$ } \\
   -1 & \text{if $x  < 0$} \\
   0 & \text{otherwise.}    
  \end{cases}
\end{equation}
The $\sgn$ operation only yields zero if its argument is exactly zero. The forward pass of course only involves multiplications with binary values.

In the hardware architecture described in the next section, we use limited precision fixed point weights during training and testing. We are  interested in quantifying the effects of limited precision weights, error ternarization, and the choice of activation function (Eq.~\ref{eq:binary_act_bipolar} or Eq.~\ref{eq:binary_act_unipolar}) on the network performance. Throughout this paper, we always use networks with two hidden layers and 600 neurons in each layer. The MNIST dataset contains 70,000 28$\times$28 grayscale images of handwritten digits~\cite{LeCun_etal98}. The training set of 60,000 labeled digits was used for training, and testing was done using the remaining 10,000 digits. All grayscale images were first binarized to two intensity values. The 784 neurons in the input layer thus have a binary output. In all trials in this section we used standard stochastic gradient descent with a mini-batch size of 100.

The only hyper-parameter we tuned was the hinge hyper-parameter $H$ used in the evaluation of the L1 loss (Eq.~\ref{eq:hinge}). We trained on 50,000 training examples while varying $H$ and chose the value of $H$ that minimized the error on the held-out remaining 10,000 training examples. We observed that this optimal value of $H$ did not depend on the activation function used (bipolar activation or unipolar activation), so we kept it fixed in all experiments. No knowledge of the test set was thus allowed to contaminate the hyper-parameter choice~\cite{Nowotny_14}. The held-out set of 10,000 training examples was then added back to the training set. In the rest of the paper, we compare the performance of different networks configurations and learning methods to find the configurations with best accuracy and lowest memory and computational overhead. We use the test set error as a comparison metric. This should not be construed as allowing knowledge of the test set to influence the network configuration, as our goal is not to set a new accuracy record but to contrast the performance of different network configurations.

Before using binary networks, we first establish an accuracy baseline for conventional ANNs using Rectified Linear Units (ReLUs)~\cite{Nair_Hinton10} and 32-bit floating-point weights. We train such a conventional ANN with two hidden layers and 600 neurons in each hidden layer on the binarized MNIST training set. We used standard stochastic gradient descent and the L1 loss and apply dropout between all layers to reduce overfitting. The conventional ANN achieves a test set error of $1.31\pm 0.04 \%$ (mean and standard deviation from 20 training trials). We retrained the same network in 20 trials using real-valued MNIST images (i.e, without binarization) and obtained a test set error of $1.11 \pm 0.05$. The binarization of MNIST digits thus hurts accuracy. However, in order to maintain a multiplier-free design, we use binarized input images throughout in our binary state networks. 

\begin{figure}[h]
\centering
  \centering
  \begin{subfigure}[b]{0.49\textwidth}
    \includegraphics[width=\textwidth]{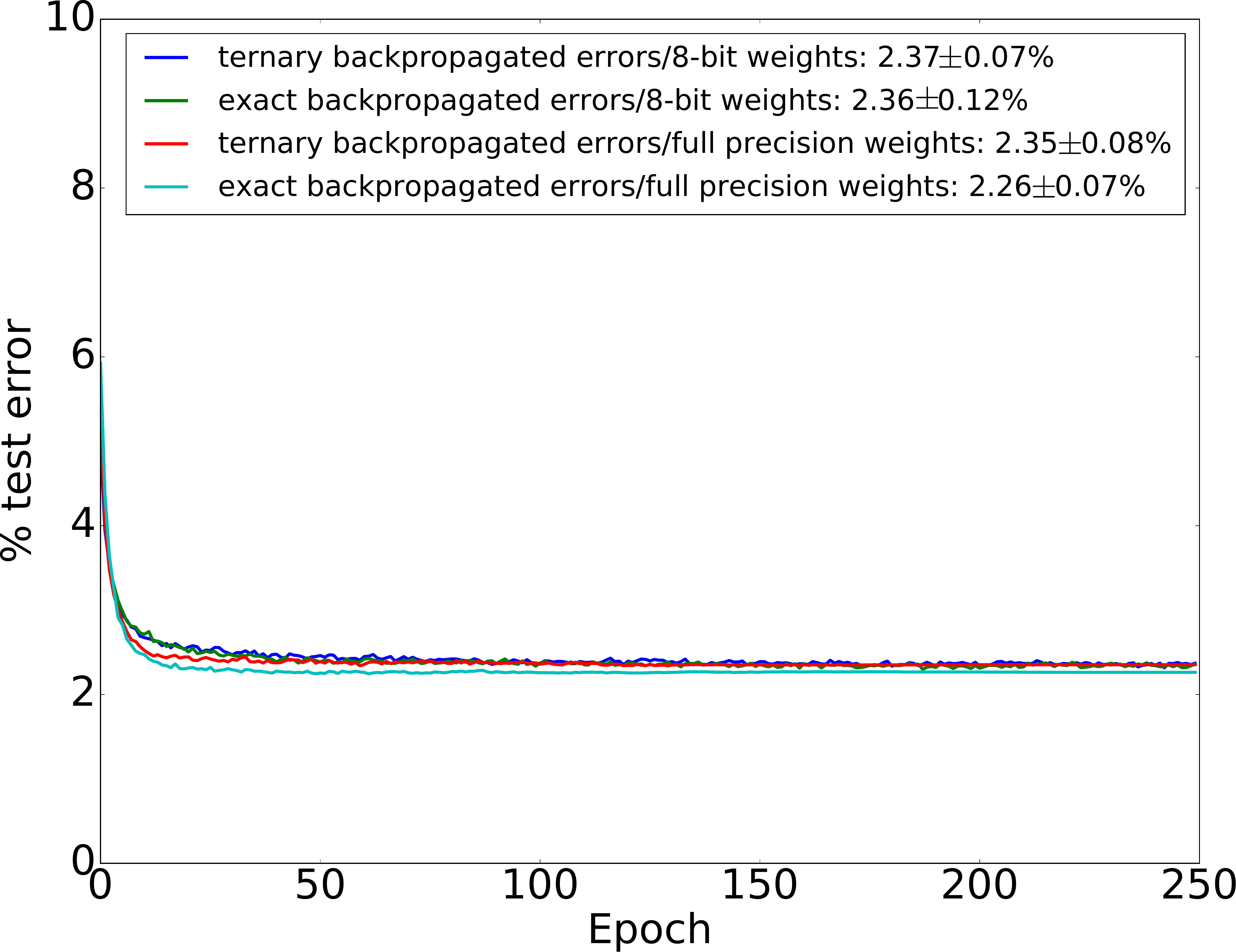}
    \subcaption{}
    \label{fig:8bits_no_dropout}
  \end{subfigure}
  \begin{subfigure}[b]{0.49\textwidth}
    \includegraphics[width=\textwidth]{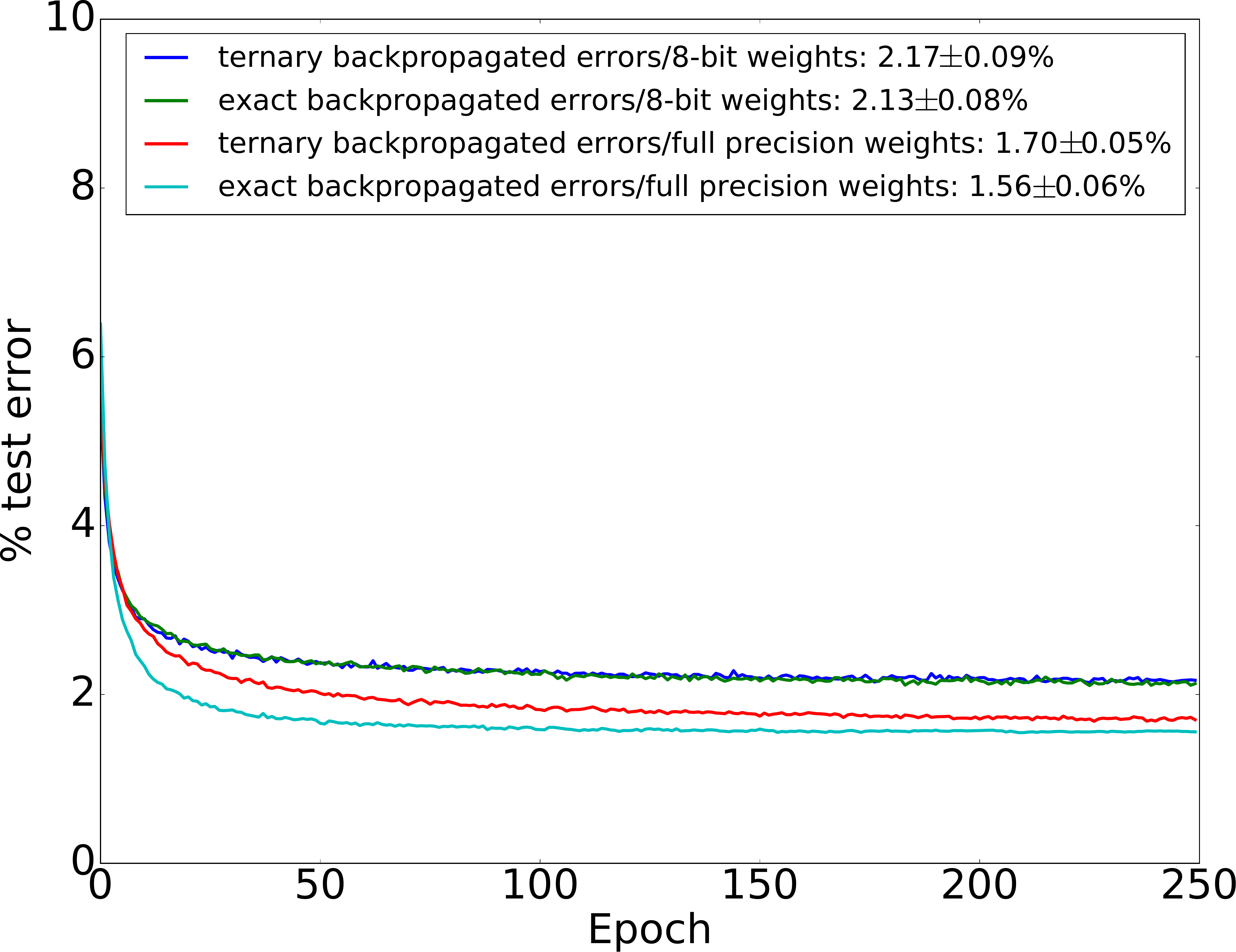}
    \subcaption{}
    \label{fig:8bits_dropout}
  \end{subfigure}
  \\
  \begin{subfigure}[b]{0.49\textwidth}
    \includegraphics[width=\textwidth]{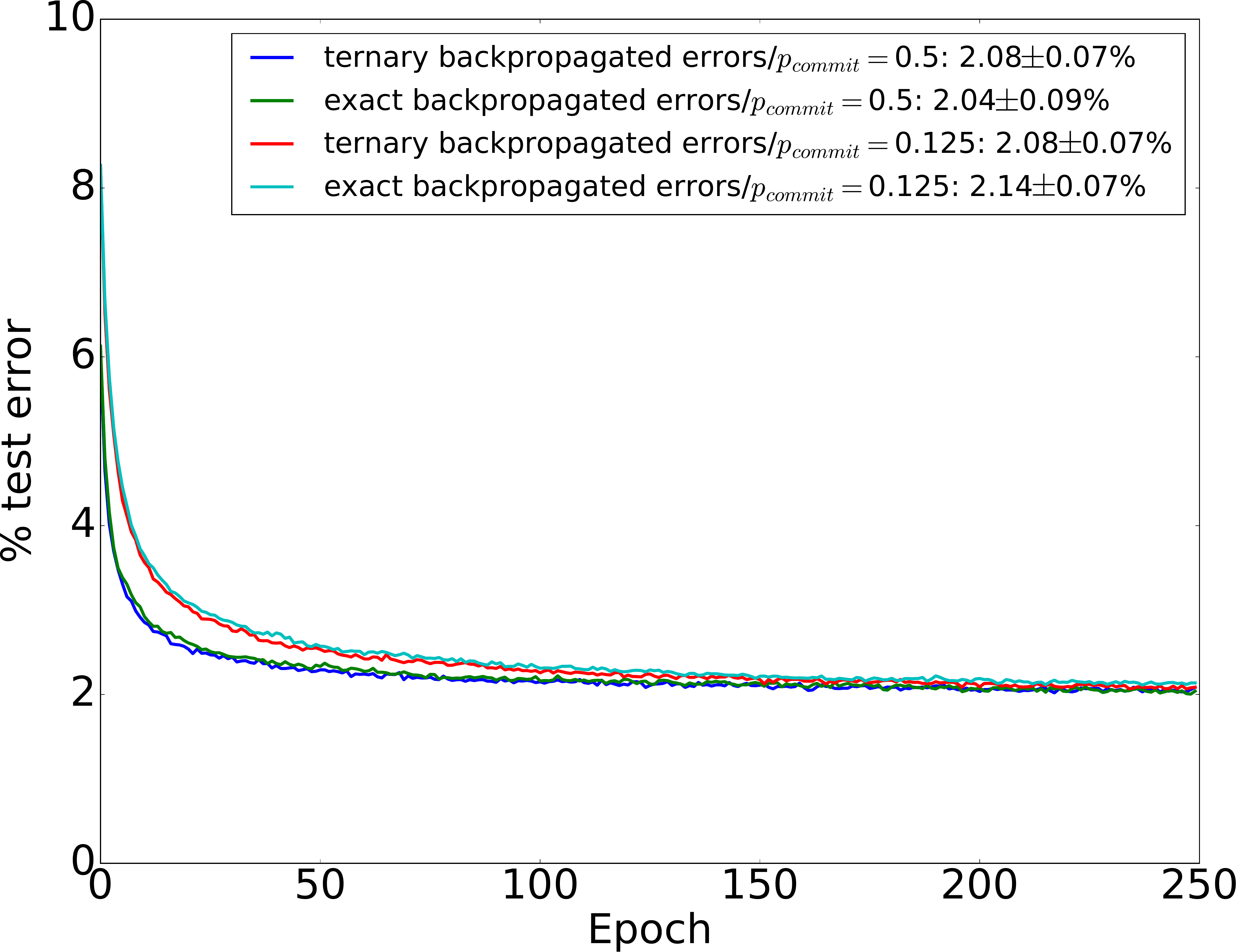}
    \subcaption{}
    \label{fig:8bits_dropout_stochastic}
  \end{subfigure}
  \begin{subfigure}[b]{0.49\textwidth}
    \includegraphics[width=\textwidth]{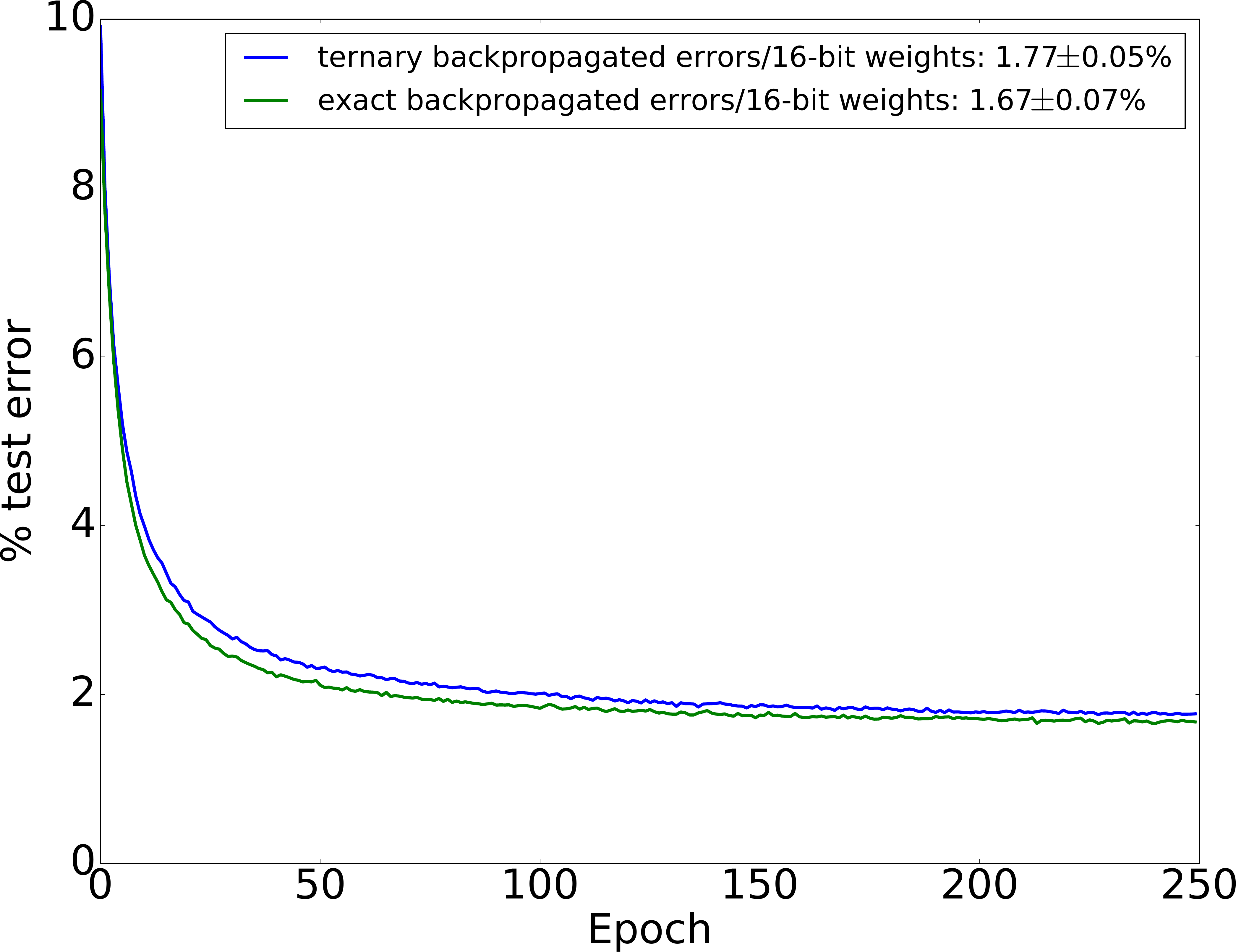}
    \subcaption{}
    \label{fig:16bits_dropout}
  \end{subfigure}

  \caption{MNIST test set errors during 250 training epochs in a network with two hidden layers and binary $-1/1$ activations. Each hidden layer has 600 neurons. Each line in the plots is an average across 20 training trials. Error figures in the legend are the final test error figures after epoch 250, together with the test error standard deviation across the 20 training trials. (\subref{fig:8bits_no_dropout}) Test errors for the four combinations of exact/ternarized backpropagated errors and 8-bit/32-bit (high precision) weights. No dropout was used. (\subref{fig:8bits_dropout}) Same as \subref{fig:8bits_no_dropout} but using a dropout probability of 0.2 between all layers during training. (\subref{fig:8bits_dropout_stochastic}) Networks trained using 8-bit weights, dropout, and  stochastic weight updates for two different values of weight commit probability, $p_{commit}$. Results for ternarized and exact backpropagated errors are shown. (\subref{fig:16bits_dropout}) Networks trained using 16-bit fixed point weights and dropout. Results for ternarized and exact backpropagated errors are shown.} 
\label{fig:mnist_experiments}
\end{figure}

We first investigate network performance using the activation function in Eq.~\ref{eq:binary_act_bipolar}, i.e, using the $-1/1$ bipolar format. Figure~\ref{fig:8bits_no_dropout} shows the effect of error ternarization (Errors coming from the top layer are not ternarized) and limited precision (8-bit) weights on the network performance. The four lines depict the evolution of the test error in the 4 combinations of exact/ternary errors and 8-bit fixed-point/32-bit floating point weights. In all four cases, the network's error on the training set reached zero.
In the 8-bit case, we used a learning rate of $1$ which is the smallest possible learning rate. In the 32-bit (high-precision) case, we used a real-valued exponentially decaying learning rate. The networks severely overfits on the training data which explains why using low-precision weights and error ternarization barely affect test accuracy.  

To combat overfitting, we applied dropout~\cite{Srivastava_etal14} to the output of each layer (including the input layer). The training results are shown in Figure~\ref{fig:8bits_dropout}. Networks with high-precision weights clearly outperform networks with 8-bit weights once the networks are regularized using dropout. Error ternarization slightly degrades accuracy in the regularized networks. However, the performance loss is small compared to the loss incurred when switching to 8-bit weights. In the 8-bit weights case, a weight update can not be smaller than $2^{-8}$ of the full weight range. Networks with 8-bit weights thus have a large effective learning rate. Small learning rates, however, are instrumental in allowing neural networks to gradually accumulate information from the entire training set. To achieve a small effective learning rate with large minimum weight updates, we tried using stochastic weight updates where the update of each individual weight is committed to memory with a probability $p_{commit}$. Weights thus change more slowly since weight updates are stochastically discarded. Performance of the network with 8-bit weights and two different commit probabilities, $p_{commit} = 0.5$ and $p_{commit} = 0.125$, is shown in Fig.~\ref{fig:8bits_dropout_stochastic}. Stochastically discarding weight updates did not appreciably improve network performance.

In order to approach the performance of networks with 32-bit floating point weights, we turn to networks with 16-bit fixed-point weights. We used a learning rate of $16$. Using this learning rate, the size of the smallest weight update is thus $2^{-12}$ of the full weight range. The performance of networks with 16-bit weights is shown in Fig.~\ref{fig:16bits_dropout}. Networks with 16-bit weights significantly outperform networks with 8-bit weights and their performance comes very close to that of high precision networks when using networks with $-1/1$ bipolar activation functions (Eq.~\ref{eq:binary_act_bipolar}).

\begin{figure}[h]
\centering
  \centering
  \begin{subfigure}[b]{0.48\textwidth}
    \includegraphics[width=\textwidth]{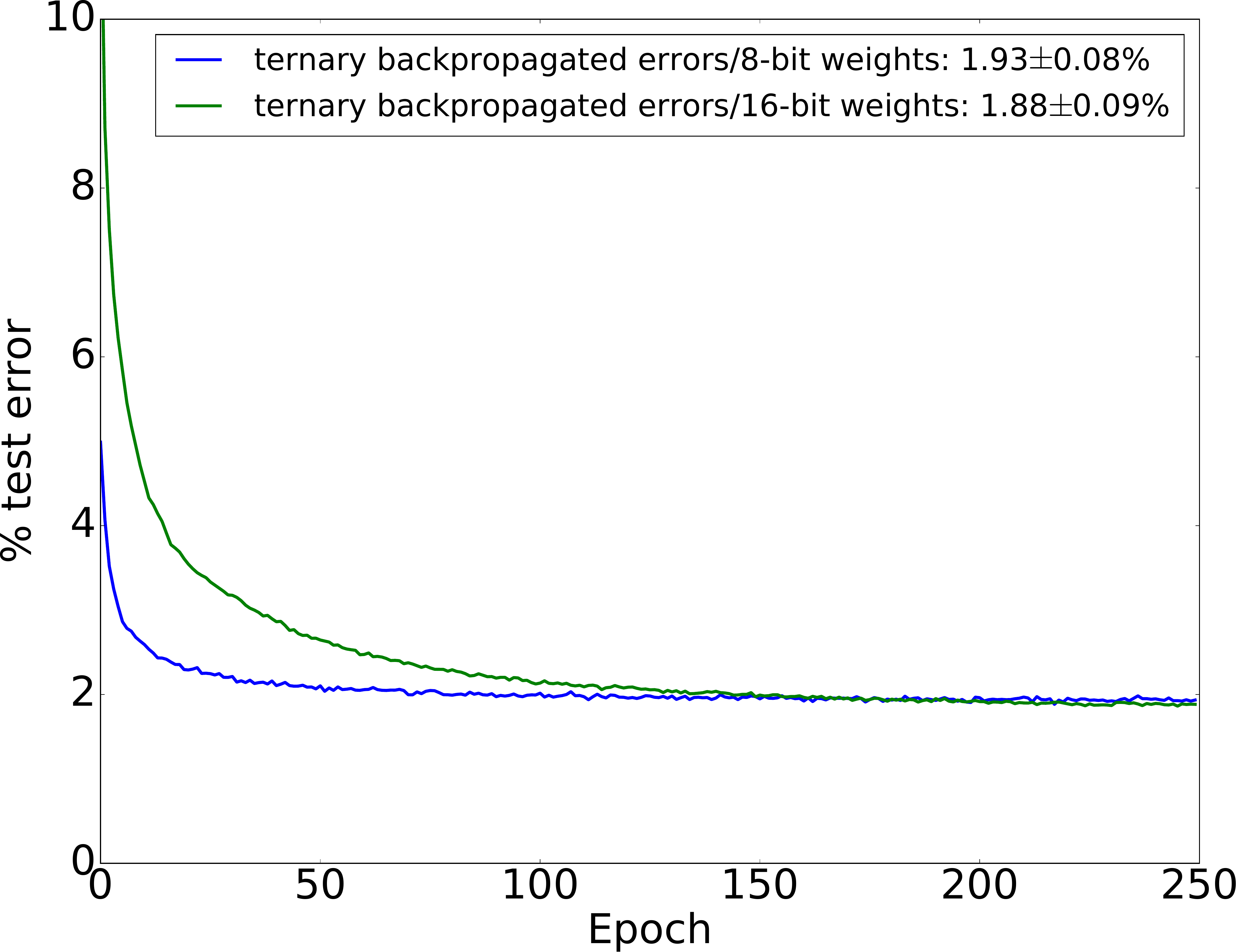}
    \subcaption{}
    \label{fig:sigmoid_accuracy}
  \end{subfigure}
  \quad
    \begin{subfigure}[b]{0.48\textwidth}
    \includegraphics[width=\textwidth]{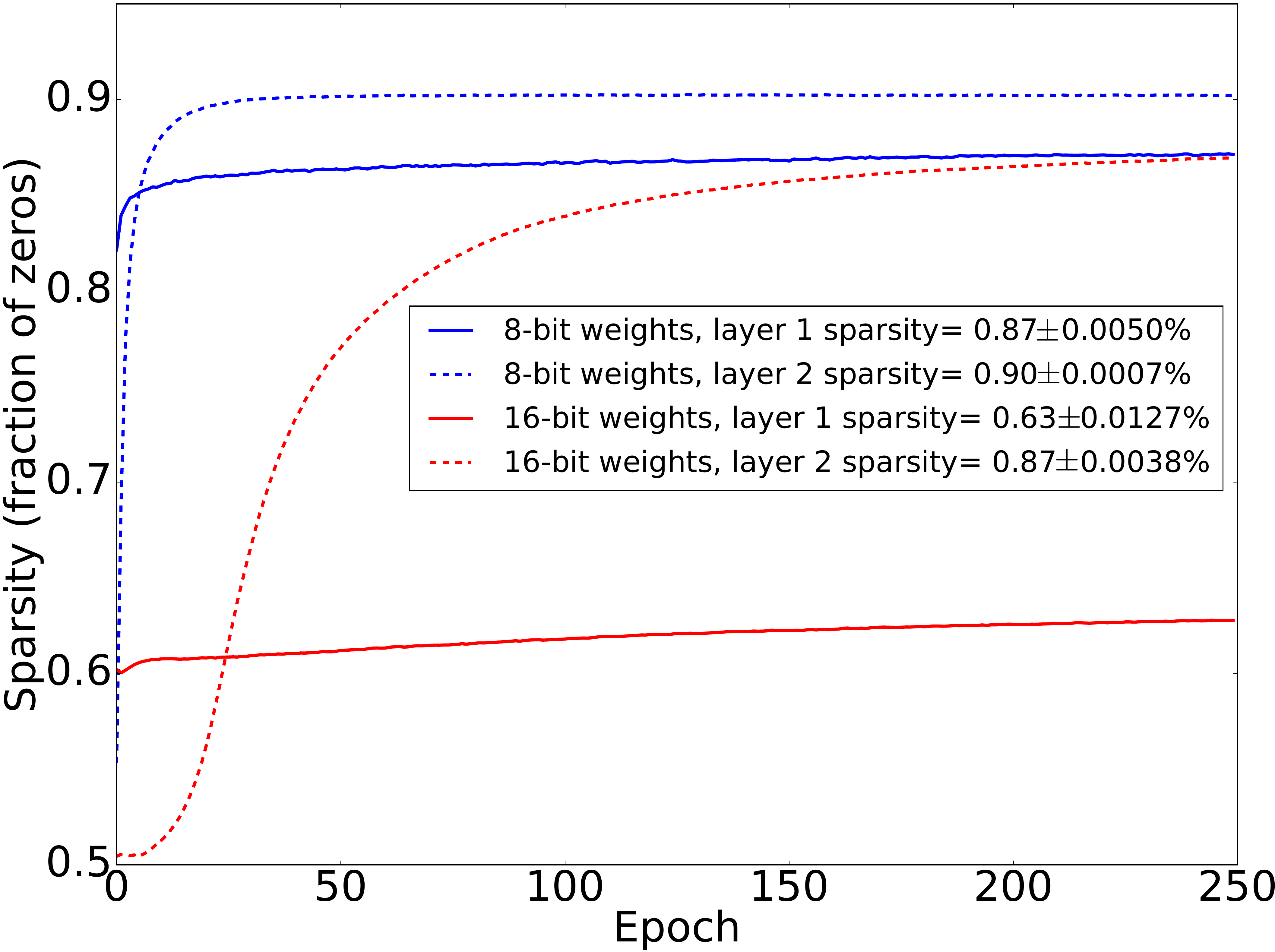}
    \subcaption{}
    \label{fig:sparsity}
  \end{subfigure}

\caption{MNIST test set errors on a similar network to the one used in Fig.~\ref{fig:mnist_experiments} except that binary unipolar $0/1$ neural activations are used instead of bipolar $-1/1$ activations. A dropout probability of 0.2 was used between all layers. Mean and standard deviations are from 20 training trials.  (\subref{fig:sigmoid_accuracy}) Test set accuracy when using ternary errors and limited precision weights (8 bits and 16 bits). Legend shows final test set error and its standard deviation. (\subref{fig:sparsity}) Sparsity (fraction of zeros ) of the activations of each of the two hidden layers when using 8-bit weights and 16-bit weights. Sparsity was evaluated on the test set after each training epoch. Sparsity figures in the legend refer to sparsity on the test set after the last training epoch. } 
\label{fig:sigmoid_activation}
\end{figure}

We switched the hidden layer activation function to the $0/1$ unipolar activation function in Eq.~\ref{eq:binary_act_unipolar} and repeated the training experiments with 8-bit and 16-bit fixed point weights. The size of the smallest weight update when using 16-bit weights is $2^{-12}$ of the full weight range, and $2^{-8}$ of the full weight range when using 8-bit weights. We applied dropout between all layers. The results are shown in Fig.~\ref{fig:sigmoid_accuracy}. The performance gap between 16-bit weights and 8-bit weights decreases significantly when using $0/1$ unipolar activations compared to $-1/1$ bipolar activations. We conjecture that this is due to the reduced number of weight updates committed during each training iteration: when using unipolar activations, many neurons will have $0$ activations which stops all their outgoing weights from being updated. This leads to slower learning compared to $-1/1$ bipolar activations where weights are always updated when there are errors coming from higher layers. Slower learning allows the network to better accumulate evidence from the entire training set. To support this conjecture, we investigated the activation sparsity in the hidden layers. The results are shown in Fig.~\ref{fig:sparsity} and they indicate the majority of neurons have $0$ activations. Sparsity increases during training and is more pronounced when using 8-bit weights. As we show in section~\ref{sec:fpga}, activation sparsity also has the beneficial effect of significantly reducing memory traffic during learning.


\FloatBarrier


\subsubsection{Pipelined backpropagation}

\begin{figure}[t]
  \centering
  \includegraphics[width=\textwidth]{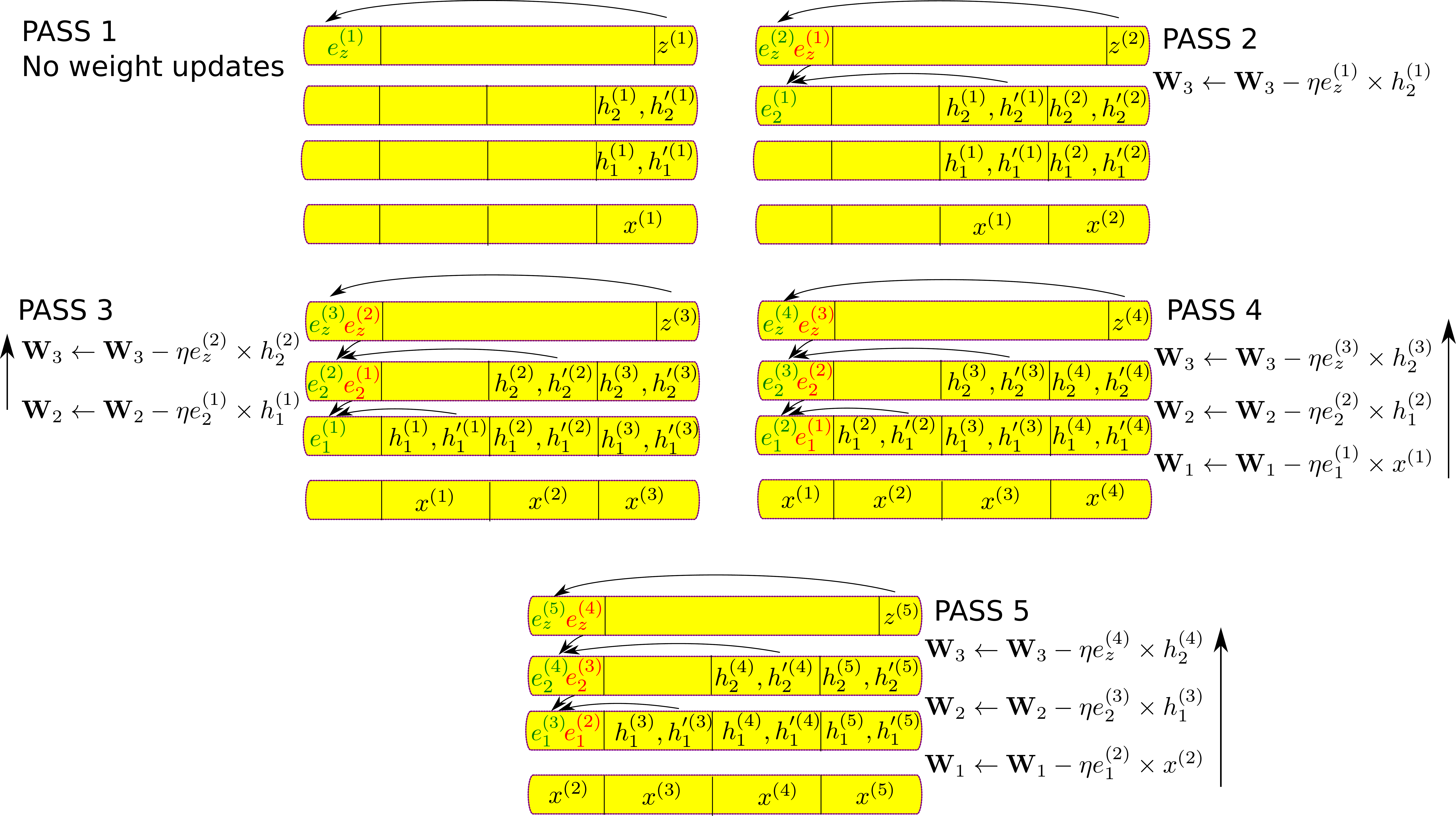} 
  \caption{Illustration of pipelined backpropagation for the two-layer network of Fig.~\ref{fig:backprop}, showing network history and storage requirements.
    The upward arrows indicate the order in which the weight updates are carried out, in the same order of weight lookups in the forward pass. Each additional layer in the network 
    incurs for each lower layer additional delay in the error computation and the weight updates, which also requires additional history of the hidden unit states to be stored.
Previous errors (shown in red) are overwritten by the newly backpropagated errors (shown in green) at the end of each pass. Hence only one error value is simultaneously stored per layer.}
\label{fig:pipelined_backprop}
\end{figure}

A straightforward implementation of backpropagation would carry out the forward pass (upward arrows in Fig.~\ref{fig:backprop}) followed by the backward pass (downward arrows in Fig.~\ref{fig:backprop}). That is because the computation in the backward pass depends on the error at the top layer and this error is only available at the end of the forward pass. Carrying out the forward and backward passes in a strict sequence, however, has the disadvantage that each network weight will usually have to be fetched twice: once during the forward pass to calculate the target neuron's activation, and once during the backward pass so that the new weight value can be calculated by incrementing/decrementing the current value before writing the new weight value to memory.


Pipelined backpropagation addresses this problem by reducing the number of redundant weight fetches and removing the need for an explicit backward pass. We reuse the notation and 2-hidden layer example from Fig.~\ref{fig:backprop}. Let $x^{(i)}$ and $target^{(i)}$ be the $i^{th}$ input pattern and target presented to the network respectively, and $h_1^{(i)}$, $h_1'^{(i)}$, $h_2^{(i)}$, $h_2'^{(i)}$, $z^{(i)}$ and $e_z^{(i)}$ be the network state after processing the input $x^{(i)}$ and $target^{(i)}$ using the latest network weights. As weights are fetched in order to propagate $x^{(i)}$ upward through the network, the network does not yet have access to the weight updates associated with input pattern $x^{(i)}$. However, it can have access to the network states associated with previously presented patterns and use these network states to carry out delayed weight updates associated with previously presented patterns.

Pipelined backpropagation is illustrated in Fig.~\ref{fig:pipelined_backprop} for the same example network from Fig.~\ref{fig:backprop}. During PASS 1, the first input and target, $x^{(1)}$ and $target^{(1)}$, are presented to the network and a complete forward pass is carried out, including the calculation of the top-level error $e_z^{(1)}$. No weight updates are carried out during PASS 1. The network maintains a history of its state during PASS 1 when processing the second input in PASS 2. During PASS 2, as the network is fetching the weights in ${\bf W}_3$ in order to calculate $z^{(2)}$, it can use the values from the previous pass $h_2^{(1)}$, $h_2'^{(1)}$, and $e_z^{(1)}$ to carry out a delayed weight update for ${\bf W}_3$ and delayed calculation of the error at the second hidden layer, $e_2^{(1)}$, from $e_z^{(1)}$. 
Similarly during PASS 3, as the network is fetching the weights in ${\bf W}_2$ the error $e_2^{(1)}$ is available and the network can update ${\bf W}_2$ as well as push down the error by one layer to obtain $e_1^{(1)}$. It is only during PASS 4 that all the weight updates associated with the first example, $x^{(1)}$ can be completed using $x^{(1)}$ and $e_1^{(1)}$. The pipeline is now full and during all subsequent passes, each weight that is fetched to compute the forward pass for the current input will also be updated using an old network state and the delayed errors that are gradually pushed down from the top layer. This is illustrated in PASS 5. Note that each layer can discard its old error (shown in red) as soon as it has enough information to calculate the new error (shown in green). That is because the old error has already been used by the preceding layer and is no longer needed. The two errors in each layer thus never have to be simultaneously stored. 

Due to the delayed weight updates, pipelined backpropagation does not yield the exact same results as standard backpropagation. For example, in PASS 3, the input $x^{(3)}$ sees the initial values of ${\bf W}_1$ and ${\bf W}_2$ but the updated value of ${\bf W}_3$. In standard backpropagation, the weights are all updated after a training example or a training minibatch. In pipelined backpropagation, the weights in higher layers are updated based on more recent input compared to the weights in deeper layers.
For large datasets such as MNIST, these slight differences in the timing of the weight updates over the course of long training epochs have negligible impact on performance in practice.

It is clear from Fig.~\ref{fig:pipelined_backprop} that pipelined backpropagation incurs extra memory overhead in order to store old network states that are needed for carrying out the delayed weight updates. 
Let $N_{inp}$, $N_{hid}$, and $N_{err}$ be the number of bits needed to store the activity of a neuron in the input layer, the activity (binary values of the activation and its derivative) of a neuron in the hidden layer, and the backpropagated error, respectively. In a network with $L$ hidden layers, the extra memory needed in the input layer to carry out the pipelined backpropagation scheme shown in Fig.~\ref{fig:pipelined_backprop} is $(L+1)\times N_{inp}$ bits per neuron. This is the extra memory compared to implementing backpropagation without pipelining. In the hidden layers, the extra memory requirements are largest in the deepest layer as this is the hidden layer which has to wait the longest to get the backpropagated error and requires $L \times N_{hid} +  N_{err}$ extra bits per neuron. The extra memory requirement per neuron successively decreases by $N_{hid}$ bits for each layer above the deepest layer.

When using smooth activation functions, the derivative of the activation value does not need to be explicitly stored as it can be inferred from the activation value itself. 
However, for BSNs, the derivative needs to be stored as it can not be inferred from the neuron's binary value. Thus 
the neuron's activity can be compactly represented as just 2 bits, 
as the neuron's output and the gradient of the activation function are both binary (Equations~\ref{eq:binary_act_bipolar},~\ref{eq:binary_act_unipolar}, and~\ref{eq:binary_der}). Since we also use dropout while training the BSN, we need an extra bit to indicate  whether the neuron was dropped in the forward pass bringing the total number of bits needed to store the neuron's state to 3: $N_{hid} = 3$, and $N_{inp} = 2$ if the input variables are also binary. Note that we do not need to store the derivative information for the input layer neurons. Implementations of conventional networks on custom hardware often use 16 bits of precision for the activation values, i.e, 16 bits are needed to store the neuron state ($N_{hid}=16$) if we assume ReLUs~\cite{Nair_Hinton10} are used. The memory overhead of implementing pipelined backpropagation in BSNs is thus $\sim$5.3x smaller compared to conventional networks. The reduction in memory overhead  becomes significant when implementing deep networks that incur a larger memory overhead to support pipelined backpropagation.

\subsubsection{Hardware Architecture}

\begin{figure}[h]
\centering
  \centering
  \begin{subfigure}[b]{0.7\textwidth}
    \includegraphics[width=\textwidth]{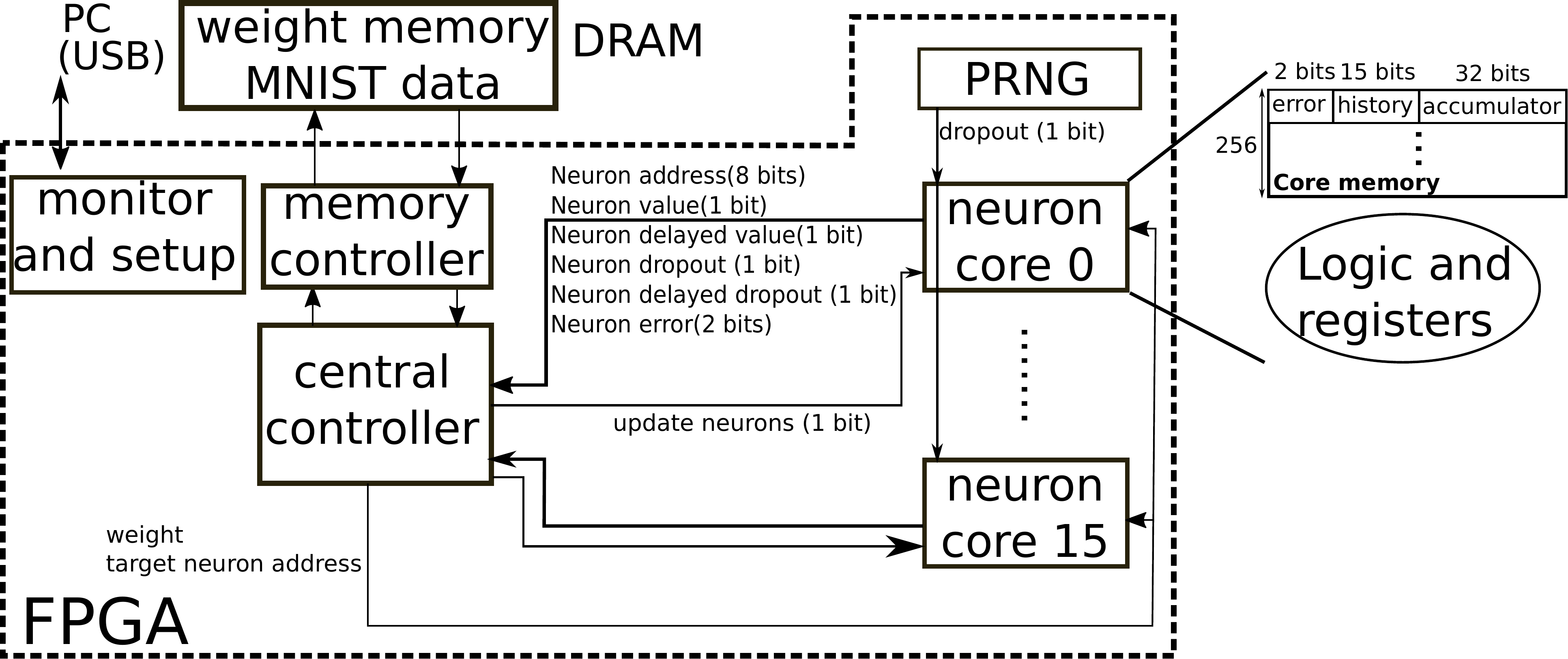}
    \subcaption{}
    \label{fig:fpga_arch}
  \end{subfigure}
  \quad
    \begin{subfigure}[b]{0.18\textwidth}
    \includegraphics[width=\textwidth]{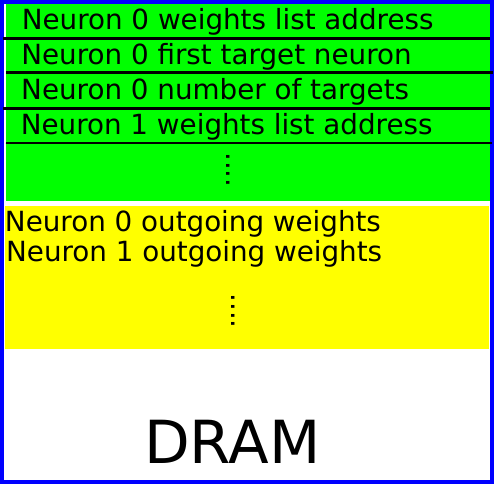}
    \subcaption{}
    \label{fig:memory_layout}
  \end{subfigure}

\caption{(\subref{fig:fpga_arch}) Block diagram of the FPGA architecture implementing pipelined backpropagation. Sixteen neuron cores with 256 neurons each are sequentially updated to realize the pipelined backpropagation scheme illustrated in Fig.~\ref{fig:pipelined_backprop}. Each core uses 256 $\times$ 49 = 12,544 bits of internal memory to store the states of the 256 neurons. A Pseudo Random Number Generator (PRNG) supplies the dropout signal to the core. The probability that the dropout signal is asserted is controlled by a configurable register in the PRNG. (\subref{fig:memory_layout}) Layout of weights in the external memory. } 
\label{fig:arch}
\end{figure}

We developed a proof-of-concept hardware architecture to illustrate the viability and classification accuracy of the proposed pipelined backpropagation scheme illustrated in Fig.~\ref{fig:pipelined_backprop}. This proof-of-concept architecture targets an FPGA platform where the weights are externally stored in DRAM. The architecture thus has very little parallelism since the central bottleneck is fetching weights from the external memory. The proposed architecture is shown in Fig.~\ref{fig:fpga_arch}. The architecture supports either 16-bit or 8-bit signed fixed-point weights and can implement neurons with $-1/1$ bipolar activations (eq.~\ref{eq:binary_act_bipolar}) or $0/1$ unipolar activations(eq.~\ref{eq:binary_act_unipolar}). The neurons are distributed across sixteen cores where each core implements 256 neurons. Each core can only contain neurons belonging to one layer. Multiple cores can be assigned to the same layer. Each core communicates with a central controller. The states of the 256 neurons in a core are stored in internal memory that is local to the core. Each neuron has a 15-bit history field divided into 5 3-bit slots which can store the neuron's state (binary output, binary derivative, and dropout state) for up to 5 passes in the past, a 32-bit accumulator field used to both accumulate forward propagating input and backward propagating error, and a 2-bit field used to store the ternary error at that neuron. Each core receives a 1-bit dropout signal from the Pseudo Random Number Generator (PRNG). The PRNG is implemented using two counter-propagating linear feedback shift registers with differing feedback length following the scheme in~\cite{Cauwenberghs96}. The dropout signal from the PRNG decides whether the currently updating neuron should be dropped for the current input. The central controller sends the update signal to each core in succession. When a core receives the update signal, it sequentially updates the states of its 256 neurons. A neuron update involves the following steps:
\begin{enumerate}
\item The neuron compares its local accumulator value to zero to decide its binary output value. It also decides the value of the binary virtual gradient by checking whether the accumulator value is in the range $[-2^{16},2^{16}]$ for 16-bit weights, or in the range $[-2^{8},2^{8}]$ for 8-bit weights  (corresponding to the range $[-1,1]$ in Eq.~\ref{eq:binary_der}). The neuron shifts the new binary value, virtual gradient, and dropout state into the 15-bit history field. The oldest 3-bit state in the history field is thus discarded. The neuron then resets the accumulator to zero. 
\item The neuron communicates its new binary output value to the controller together with its dropout state. The neuron also communicates its delayed output value and delayed dropout state from $K$ passes in the past. These delayed quantities are fetched from the history field. $K$ is different for neurons in different layers and is larger for neurons in deeper layers. $K$ is the same for neurons in the same core as each core contains neurons from the same layer so $K$ is stored in a central core register instead of in the neuron. 
\item If  the current dropout state of the source (updating) neuron is deasserted and the neuron's output is not $0$:
The controller fetches the updating neuron's outgoing weights from the external memory. For each fetched weight, it multiplies the weight by the source (updating) neuron's current binary value and dispatches the result to the target neuron. The target neuron updates its accumulator using the incoming data and outputs its current error value. 
\item If the delayed-input dropout state of the source (updating) neuron is deasserted, the controller checks the source neuron's delayed output and delayed binary gradient. If either is non-zero:
The controller fetches the updating neuron's outgoing weights from the external memory (if they had not been fetched in the previous step). The controller reads the error from the target neuron. If the error is ternary, the controller multiplies this error by the weight and sends the result to the updating (source) neuron.  The neuron accumulates the incoming value into the accumulator field. If the error is not ternary, i.e, it is coming from a top layer neuron, then it will have an absolute value of at most $C-1$. Denote this error by $e_z[i]$. Instead of multiplying the weight by the error, the controller dispatches the weight with the appropriate sign $e_z[i]$ times to the target neuron in order to implement multiplications through repeated additions. The controller multiplies the delayed output of the updating neuron (which is binary) by the delayed error from the target neuron to calculate the weight update, then writes the updated weight to memory if the weight update is non-zero.  
\item After all the outgoing weights of the updating neuron have been processed, the updating neuron fetches the binary gradient value from $K$ passes in the past from the history field. This binary gradient is then multiplied by the accumulator value (which now contains the weighted sum of the backpropagated delayed errors from all the target neurons) to obtain the new high precision error in the neuron. This error is directly ternarized according to Eq.~\ref{eq:approx_error} and stored in the error field. The accumulator is then reset.
\end{enumerate}

Note that the neuron accumulator which is used to accumulate the incoming neuron input coming from the layer below is also reused for accumulating the errors coming from the layer above. Neurons in the output layer are special. Updating these neurons is done internally in a special neuron core (core 15) based on the current input label. The core spends $C$ cycles to obtain the current classification result by finding the neuron with the maximum value among the $C$ output neurons. During these $C$ cycles, it also calculates the new top layer errors using Eq.~\ref{eq:output_update}.

The input layer neurons are implemented the same way as the hidden layer neurons. At the beginning of each pass, the accumulators of the input neurons are set to +1/-1 to encode the binary-valued input vector. This is wasteful in terms of memory resources as the extra 31 bits of the accumulator and the 2-bit error field are not needed for the input neurons but it leads to a more uniform implementation. A pass consists of setting the input neuron values then updating all the cores in succession which implements the pipelined backpropagation scheme depicted in the example in Fig.~\ref{fig:pipelined_backprop}. For the first few initial passes, the backpropagation pipeline will not be full (PASS 1 and PASS 2 in Fig.~\ref{fig:pipelined_backprop} for example). Special registers in the controller handle this initial phase by only committing weight updates calculated using valid error and delayed neuron output values. Since the network has to be updated from bottom to top (see Fig.~\ref{fig:pipelined_backprop}), neurons belonging to one layer should occupy a core with a lower index than neurons belonging to a higher layer.

Each core has a flag that indicates how the binary value of the neuron should be interpreted: either as $-1/1$ or $0/1$. This flag is communicated to the controller and influences how the target neurons and the weights are updated. In the $0/1$ interpretation, no weight is sent to the target neuron if the updating (source) neuron value is low (as opposed to sending the negative of the weight in the $-1/1$ interpretation). Moreover no weight update is carried out if the delayed neuron value is low as multiplying this delayed value by the target neuron's error would yield zero. The virtual gradient calculated using Eq.~\ref{eq:binary_der} is independent of the binary interpretation. 

Figure~\ref{fig:memory_layout} shows the structure of the data in the external memory. To fetch a neuron's outgoing weights and targets, the neuron's address is used as an index in the green region to fetch the location of the neuron's outgoing weights list, and to fetch the number of target neurons and the address of the first target neuron. Since a neuron always targets a consecutive set of neurons, only the starting neuron and the number of targeted neurons are needed. A neuron's weights list is an ordered list of weights specifying the outgoing weights to all the target neurons. Each 32-bit word in memory will contain either 4 weights or 2 weights depending on whether 8-bit weights or 16-bit weights are being used. While full-indexing in weight lookup could be used to provide greater flexibility in sparse reconfigurable synaptic connectivity~\cite{Park_etal17}, the implemented lookup scheme is more compact, incurring only a small memory overhead (the green region) when storing the weights. In our particular implementation, this overhead is 64 bits per neuron.

\section{Results}
\label{sec:fpga}
We implemented the proposed architecture on a Spartan6-LX150 FPGA. The external memory is a DDR2 memory. Each core stores the states of its 256 neurons in two 9-kb block RAMs. Due to the little parallelism in the architecture, the FPGA implementation takes up a small fraction of the FPGA resources. A breakdown of the FPGA resource utilization per block is shown in Table~\ref{table:resources}. We ran the FPGA core (which does not include the DDR2 memory controller or the USB interface) at a frequency of $78$\,MHz. The critical path occurs between the block RAM output data pins in one neuron core and the block RAM input address registers in another. This critical path is active during the backpropagation of the ternary error from a target neuron to a source neuron. The ternary error affects the address pins as it determines whether the error accumulator in the source neuron needs to be updated (if the ternary error is non-zero) or not. We configured the FPGA to implement a 2-hidden layer network with 600 neurons in each hidden layer, and to train the network on the MNIST dataset. The input layer of 784 neurons occupied 4 neuron cores and was configured as a unipolar $0/1$ layer. Each hidden layer occupied 3 neuron cores and was configured either as a $-1/1$ bipolar layer or $0/1$ unipolar layer. The output layer was implemented on a special core (core 15) designed to calculate the error based on the hinge loss from Eq.~\ref{eq:hinge}. The weights in the DDR2 memory were initialized using the initialization scheme in~\cite{Glorot_Bengio10} . The MNIST images were binarized and stored in the DDR2 memory together with their labels. Each MNIST image/label pair takes up $784 + 4=788$ bits. The controller fetches the training images/labels sequentially from memory to configure the input layer (using the pixel values) and the output layer (using the label). Evaluation of the test set is also done on FPGA after switching off learning.

\begin{figure}[t]
  \centering
  \begin{subfigure}[b]{0.46\textwidth}
    \centering
    \includegraphics[width=1.0\textwidth]{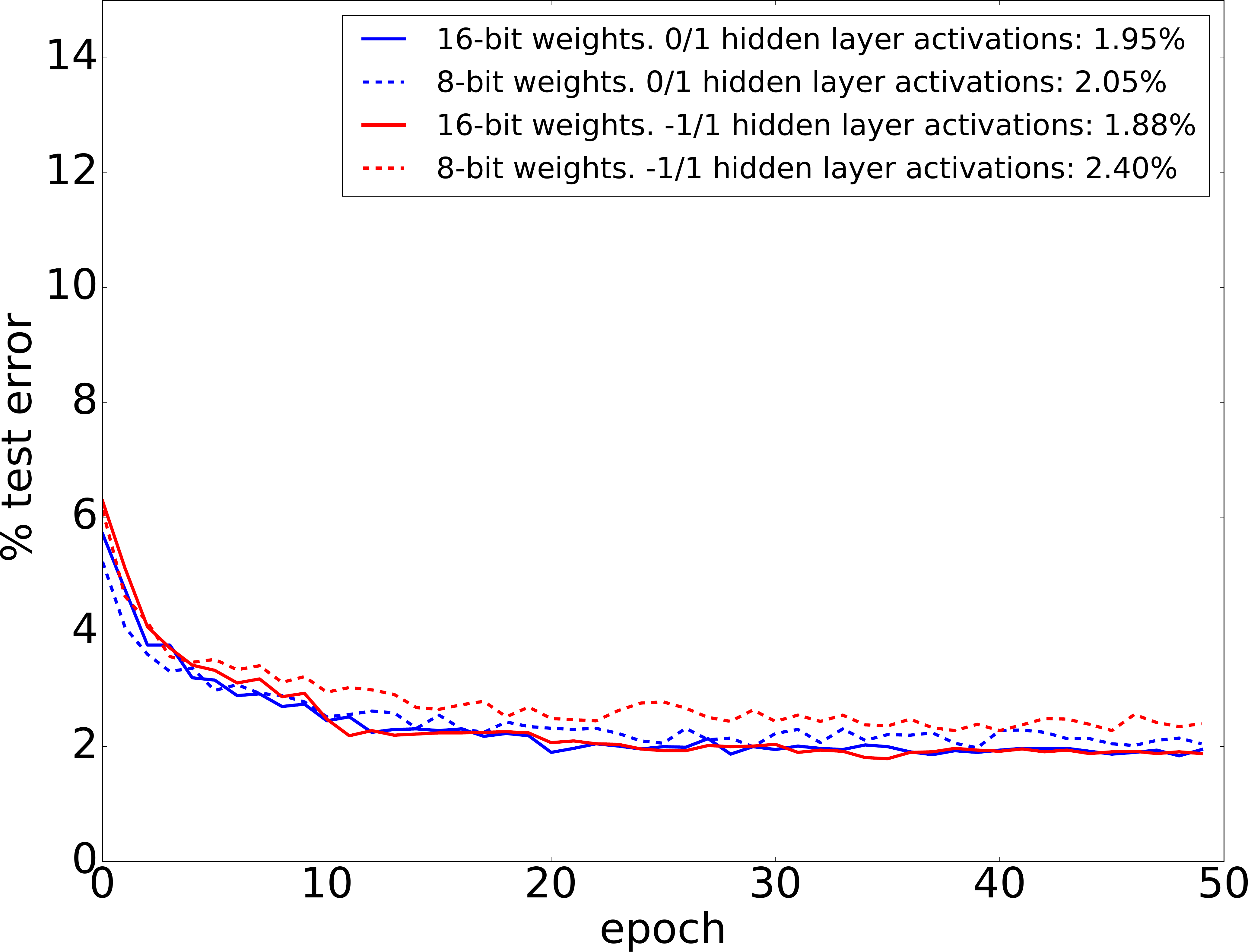}
    \subcaption{}
    \label{fig:mnist_fpga}
  \end{subfigure}
  \quad
  \begin{subfigure}[b]{0.46\textwidth}
    \centering
    \includegraphics[width=1.0\textwidth]{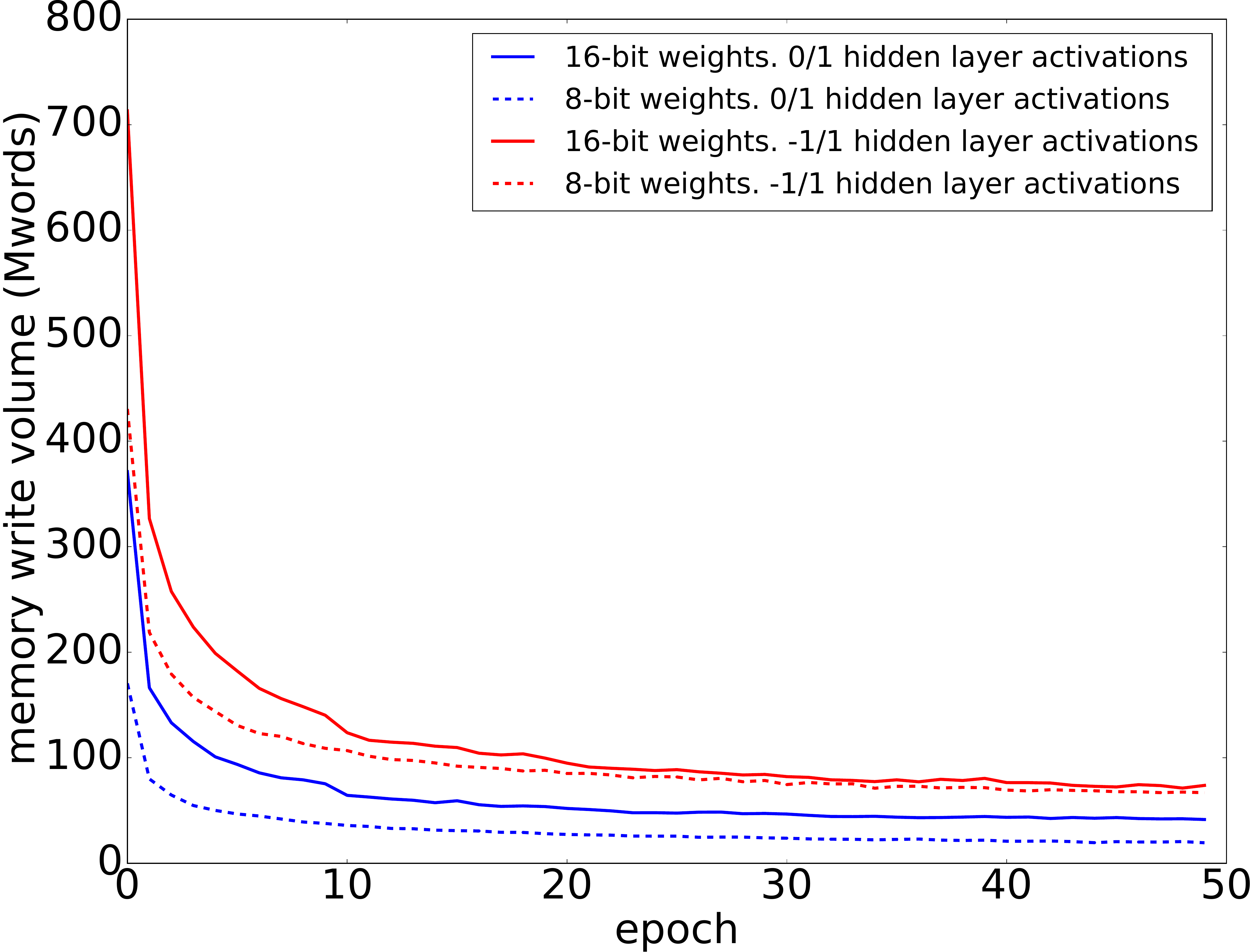} 
    \subcaption{}
    \label{fig:write_volume}
  \end{subfigure}
  \\
  \begin{subfigure}[b]{0.46\textwidth}
    \centering
    \includegraphics[width=1.0\textwidth]{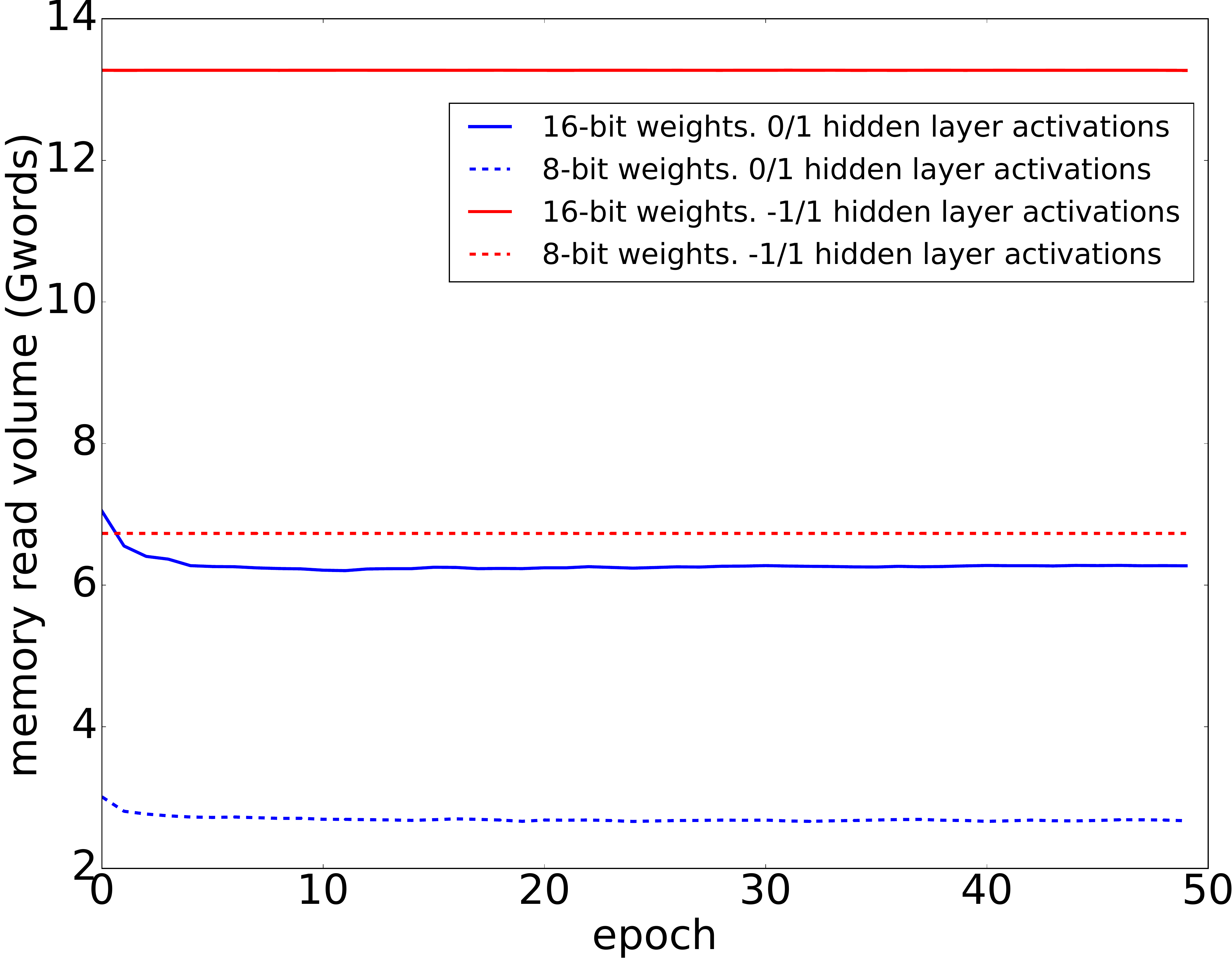}
    \subcaption{}
    \label{fig:read_volume_words}
  \end{subfigure}
  \quad
  \begin{subfigure}[b]{0.46\textwidth}
    \centering
    \includegraphics[width=1.0\textwidth]{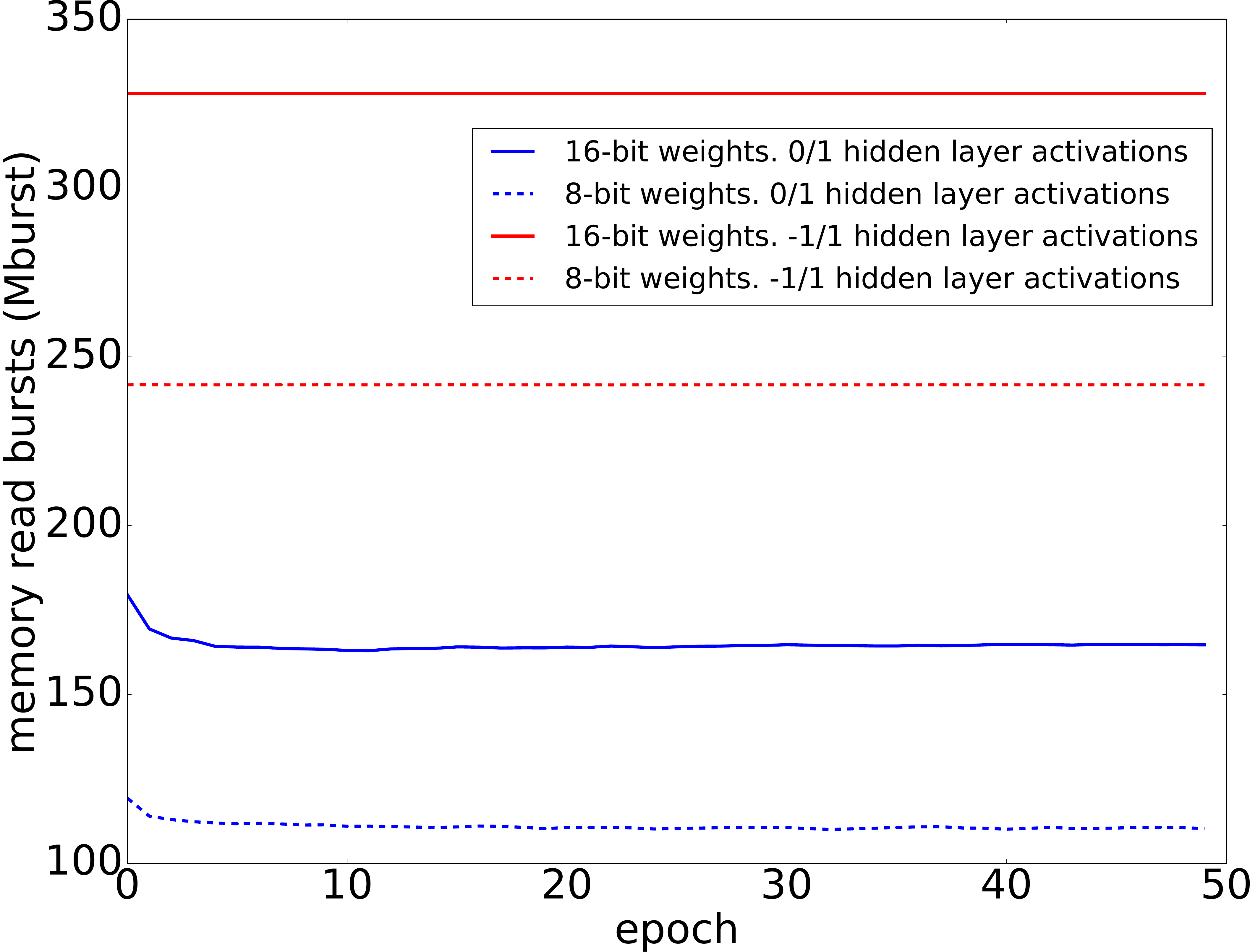} 
    \subcaption{}
    \label{fig:read_volume_bursts}
  \end{subfigure}

  \caption{Test-set error and memory access statistics when training four different network configurations corresponding to the four combinations of 8-bit/16-bit weights and unipolar/bipolar activations for 50 epochs. Training and testing were conducted on the FPGA using a 784-600-600-10 network. (\subref{fig:mnist_fpga}) Test set errors on the MNIST dataset. The error figures in the legend refer to the final error figures after epoch 50. (\subref{fig:write_volume}) Number of 32-bit words written to DDR2 memory during each training epoch. (\subref{fig:read_volume_words}) Number of 32-bit words read from DDR2 memory during each training epoch. (\subref{fig:read_volume_words}) Number of read bursts during each training epoch. Each burst can be up to 64 words long.}
\label{fig:fpga_results}
\end{figure}

The FPGA trained on MNIST training set digits for 50 epochs. We trained using four different network configurations corresponding to the four combinations of 8-bit/16-bit weights and unipolar/bipolar activations. The FPGA implements a dropout probability of 0.2 between all layers in all network configurations. When using 8-bit weights, the weight update magnitude is $1$. When using 16-bit weights, a weight update magnitude of $1$ ($2^{-16}$ of the full weight range) results in very slow learning. For 16-bit weights, we instead use a learning rate (update magnitude) adjustment scheme that is analogous to the exponentially decaying learning rate schemes used to train conventional ANNs with floating point weights: we start with an update magnitude of $128$ and halve this update magnitude every 10 epochs. The test errors after each training epoch are shown in Fig.~\ref{fig:mnist_fpga} for the four network configurations. The results in Fig.~\ref{fig:mnist_fpga} are consistent with the accuracy figures obtained using mini-batch training and standard backpropagation (Figs.~\ref{fig:mnist_experiments} and~\ref{fig:sigmoid_activation}) where bipolar activations slightly outperform unipolar activations when using 16-bit weights. When using 8-bit weights, the situation is reversed with bipolar activations resulting in worse performance. 

\begin{table}[t]
  \caption{Register and LUT resources needed for the implementation of the architecture in Fig.~\ref{fig:fpga_arch} \\on Spartan6-LX150 FPGA. The percentage utilization for each resource is shown in the bottom row. }
  \label{table:resources}
  \centering
  \begin{tabular}[t]{lllp{35mm}}
    \toprule
    Block name     & Registers  & LUTs & 18-kb block RAM elements     \\
    \midrule
    Neuron core/top layer ($\times$1)  &  520  & 915 & 0     \\
    Neuron core/input and hidden layer ($\times$15)  & 35${\times}$15   & 225$\times$15  & 1$\times$15     \\        
    PRNG ($\times$1) & 58   & 19  & 0     \\        
    Central controller ($\times$1)  & 627    & 1413  & 0      \\
    DDR2 memory controller ($\times$1)            & 263   & 411  & 0      \\
    USB monitor and setup ($\times$1)  & 2778   & 2999  & 1     \\
    \midrule
    {\bf Total} & 4771 (1$\%$)   & 9123 (3$\%$)  & 16 (6$\%$)   \\        
    \bottomrule
  \end{tabular}
\end{table}

We inserted monitoring logic into the FPGA to count the number of DDR2 memory read and write operations while training. Figure~\ref{fig:write_volume} shows the number of 32-bit words written to memory during each training epoch. Note that the number of updated weights can not be exactly inferred from this plot as each 32-bit word can contain either 4 weights or 2 weights, and a full 32-bit word will be written to memory whenever one or more of the weights it contains have been updated. In all configurations, the volume of words written to memory per epoch drops as training proceeds; as the network makes fewer mistakes, fewer errors are generated by the top layer and fewer weight updates are performed. The use of unipolar $0/1$ activations results in sparser weight updates and less weight write volume. Even when using 16-bit weights, unipolar activations result in less weight write volume compared to using 8-bit weights with bipolar activations.

Figure~\ref{fig:read_volume_words} shows the number of 32-bit words read from memory during each training epoch. In addition to the network weights, this read volume includes the overhead needed to read a source neuron's target address range and the location of its weight table in memory (This is the data in the green region in Fig.~\ref{fig:memory_layout}). For bipolar activations, this read volume does not change during training as the outgoing weights for each hidden layer neuron always need to be fetched for the forward pass. For unipolar activations, this read volume drops slightly at the beginning as activity in the hidden layers becomes sparser and no weights are fetched for neurons with $0$ current output and $0$ delayed output. This is consistent with the increase in sparsity during training which is observed in Fig.~\ref{fig:sparsity}. Since a neuron's outgoing weights are stored in contiguous memory positions, this allows us to access the memory more efficiently by using long read bursts to read these outgoing weights. The maximum read burst size is 64 words. Figure~\ref{fig:read_volume_bursts} shows the number of read bursts during each training epoch. It is clear unipolar activations result in significantly less read traffic. Unipolar activations with 8-bit weights result in the smallest DDR2 read/write volume.

\begin{table}[h]
  \caption{Performance metrics for four different network configurations, and reductions in memory read volume realized by pipelined backpropagation }
  \label{table:performance}
  \centering
  \begin{tabular}[t]{lp{15mm}p{15mm}p{35mm}p{20mm}}
    \toprule
    Network configuration     & Write volume (Gwords) & Read volume (Gwords)  & Percentage reduction in read volume due to pipelined backpropagation & Average training time per example (ms)      \\
    \midrule
    16-bit weights. 0/1 hidden activations  & 3.24   & 314  & 15\%  & 15.0    \\
    8-bit weights. 0/1 hidden activations  & 1.63   & 135  & 12\%  & 12.0    \\
    16-bit weights. -1/1 hidden activations  & 6.03   & 663  & 36\%  & 32.8    \\
    8-bit weights. -1/1 hidden activations  & 4.90   & 337  & 36\%  & 32.5    \\    
  \end{tabular}
\end{table}

\begin{figure}[h]

  \centering
  \includegraphics[width=1.0\textwidth]{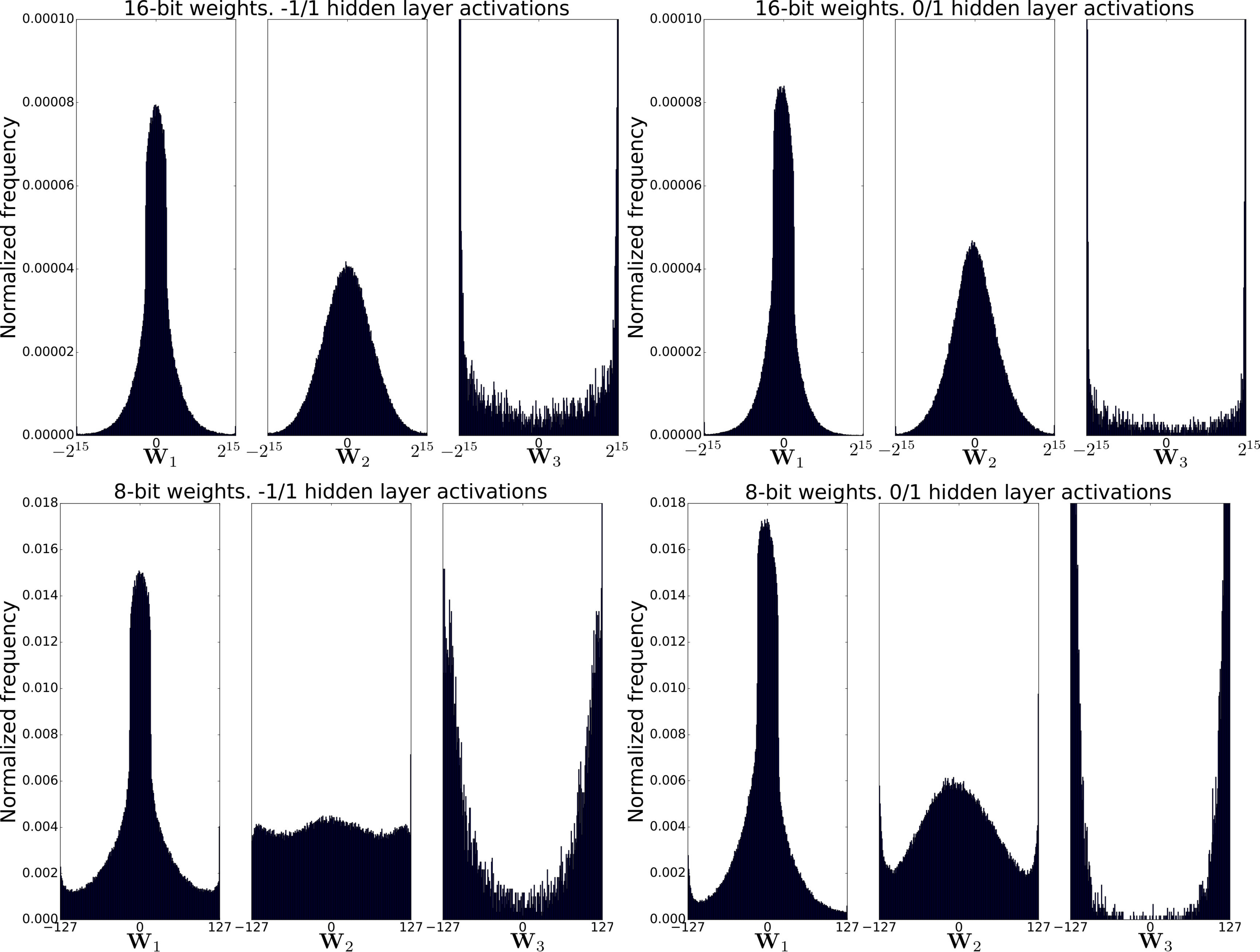} 
  
  \caption{Distribution of weights after training for four different network configurations. Significant weight clipping is observed for the weights between the second hidden layer and the output layer in all configurations. Weight clipping is more pronounced for 8-bit weights.}
\label{fig:weight_distro}
\end{figure}

All the results shown in Fig.~\ref{fig:fpga_results} were obtained when the FPGA was implementing the pipelined backpropagation scheme in which the forward and backward passes are carried out simultaneously. In order to quantify the reduction in read memory traffic due to the use of pipelined backpropagation, we inserted extra logic into the central controller to predict the memory traffic that would arise if pipelined backpropagation was not used. This prediction is straightforward as the controller has access to the updating neuron's current state and delayed state and can use this information to predict whether standard backpropagation would have needed to look up the same weight twice when training one example, once in the forward pass and once in the backward pass. Standard backpropagation might not need to look up a weight in the forward pass (if the source neuron's value is $0$ or if it has been dropped out), or it might not need to look up the weight in the backward pass (if the source neuron's value is zero and its activation derivative is zero so there is no weight update and errors can not backpropagate through the neuron). The prediction logic takes all these situations into account when predicting standard backpropagation read memory traffic. Table~\ref{table:performance} summarizes the total DDR2 read/write volume summed across all 50 training epochs, in addition to the reduction in read traffic realized by pipelined backpropagation and the average training time per example. This training time is the time needed for a full forward pass interleaved with a delayed backward pass and weight updates. The training time figures in the table were obtained by dividing the total training time, comprising $50$ epochs and $60,000$ examples per epoch, by $50\times 60,000$.

As shown in Table~\ref{table:performance}, even though unsigned binary 0/1 activations lead to overall lowest read volumes and shortest training times, the reduction in memory read volume due to pipelined backpropagation is more significant when using bipolar binary -1/+1 activations. That is because in the case of bipolar activations, it is more likely that a weight fetched to execute the forward (backward) pass will also be needed for the backward (forward) pass of the previous/delayed (later) input, allowing pipelined backpropagation to reduce read access by fetching the weight once compared to standard backpropagation which would have needed to fetch the weight twice. The reduction is not 50\% for bipolar activations due to the use of dropout which sometimes obviates the need to look up a weight in the forward and backward passes. Moreover, we always use a unipolar input layer; in some cases, the current value of an input layer pixel can be $1$ and the delayed value $0$. In pipelined backpropagation, this pixel's outgoing weights would need to be looked up to execute the current forward pass (since the pixel's value is $1$) but they will not be used in the backward pass since no weight updates are needed (since the pixel's delayed value is zero). In such situations, standard backpropagation would also need to look up the pixel's outgoing weight only once for the current example (since these weights are not needed for the forward or backward pass of the delayed example).


Figure~\ref{fig:weight_distro} shows the post-training distribution of weights in each of the three weight matrices in the network and for each of the four network configurations. As expected, weight clipping is more apparent when using 8-bit weights. The distribution of the output weights (${\bf W}_3$) is markedly more skewed towards the weight limits compared to the other two weight matrices. This could be due to the fact that these weights are the only weights that can be incremented/decremented by more than the learning rate for each training example because they see the non-ternarized error from the top layer whose magnitude could be as large as $C-1=9$. Weights in the other two weight matrices always see a ternarized backpropagated error so they can not change by more than the learning rate for each training example.

\section{Conclusions and Discussion}
\label{sec:conclusions}
We presented a scheme for the efficient implementation of pipelined backpropagation to train multi-layer feedforward networks. Due to the use of binary-state networks, the scheme is highly efficient in terms of logic and memory resources. We developed a proof-of-concept hardware architecture and implemented the architecture on FPGA to validate the proposed approach for pipelined training of BSNs. Due to error ternarization, the core operation in the forward and backward passes is fixed-point addition/subtraction. This is an equivalent operation to the synaptic operation (SynOp) which is the basic operation in spiking neuromorphic systems. SynOps can be up to two orders of magnitude more efficient than conventional MAC operations~\cite{Merolla_etal14c,Neftci_etal15}. Perhaps the biggest advantage of pipelined backpropagation is that it reduces the number of weight fetches compared to sequential forward and backward passes. In state of the art networks where the weights are too many to fit into the accelerator memory, reducing the off-chip weight traffic can lead to significant energy and performance gains. Moreover, no reverse lookup of weights is needed and weights can be stored in a way that only optimizes lookups using the source neuron address.

Pipelined backpropagation enables layer-level parallelism. A layer can begin processing a new input vector from the layer below as soon as it has finished processing the previous input vector and fetching the weighted errors from the layer above. We did not implement layer-level parallelism as it would require each layer to have access to its own weight memory in order to operate independently from the other layers. One of the main shortcomings of pipelined backpropagation is the extra memory needed to store the delayed errors and network state. Through the use of binary activation functions in the forward inference pass and ternary errors in the backward learning pass, this extra memory is kept to a minimum which makes pipelined backpropagation a feasible option for training deep networks.

The performance on the MNIST dataset was adequate but not state of the art. A natural extension of the presented architecture and algorithm would be the implementation of convolutional feedforward networks which achieve superior performance on learning tasks with a spatial structure such as vision-related tasks. When training BSNs,or binarized neural networks~\cite{Hubara_etal16}, normalization techniques such as batch-normalization~\cite{Ioffe_Szegedy15} are used to center and normalize the variance of the input to each neuron. This is particularly important for BSNs due to the hard non-linearity used and the fact that the gradient only flows back when the input to the neuron is around zero. This is clearly useful when learning static datasets using mini-batches. In an online setting with continuous learning (effective minibatch size of 1) and continuously changing inputs, it is unclear how normalization should be applied. An online normalization technique has to take into account that the input statistics could change quickly during online learning in a real-world environment, hence a normalization technique with weak history dependence is preferred~\cite{Ba_etal16}. A hardware-efficient normalization technique that can be applied online is thus a clear next step to allow the proposed architecture to train deep networks online. 


Spiking neural networks are an alternative network paradigm that is similar in many respects to BSNs in terms of required hardware resources; both types of networks can be built using adders and comparators and require no multipliers. Spiking networks have been used to solve various classification tasks~\cite{OConnor_etal13,Diehl_etal15,Cao_etal15,Lee_etal16,Esser_etal16}. Even though they require very similar computational resources, the energy, memory access patterns, and time needed to carry out the inference/forward pass in spiking networks and BSNs can be significantly different. Spiking networks are often used in the rate-based mode where the output value of a spiking neuron is encoded in its average firing rate. Multiple lookups of the same weights are thus needed to dispatch multiple spikes from the same neuron which could significantly raise energy consumption. The weight lookups are also more irregular compared to BSNs since neurons spike in an asynchronous manner. This reduces the ability to pipeline memory accesses. Spiking networks are dynamical systems which are emulated using time-stepped dynamics in digital implementations~\cite{Merolla_etal14c,Khan_etal08} or using native analog dynamics~\cite{Benjamin_etal14,Park_etal14,Qiao_etal15,Schemmel_etal10} in analog/mixed-signal implementations. The dynamic nature of spiking networks results in an irregular computational load as there could be intervals where the network is quiescent and intervals where many neurons spike simultaneously or in rapid succession. This makes it difficult to consistently achieve optimal utilization of computational and memory access resources, unlike BSNs where data movement and computations are much more predictable.

The computational and weight lookup overhead in a BSN using bipolar activations is roughly equivalent to that of a spiking network where each neuron spikes exactly once. When using unipolar activations, we observe that hidden layer activity becomes quite sparse (see Fig.~\ref{fig:sparsity}) which is reflected in the greatly reduced memory traffic when using unipolar activations instead of bipolar activations. To match the low memory traffic of unipolar BSNs, a spiking network would need to have sparse activity where $80-90 \%$ of the neurons do not spike (see Fig.~\ref{fig:sparsity}) for each classification decision.
We thus obtain sparsity-induced power savings in the synchronous setting as used with ReLU activations~\cite{Han_etal16} with BSNs having unipolar ($0,1$) rather than bipolar ($-1,+1$) activations.
Spiking networks, however, have a decisive advantage when processing dynamic and sparse event-based data such as the event trains coming from neuromorphic sensors~\cite{Lichtsteiner_etal06b,Liu_etal10,Liu_Delbruck10} as the networks can scale their spiking activity in response to the dynamically changing input event stream. The spiking network could thus effectively shut down during intervals when there is no input activity, saving power.


One of the main advantages of BSNs is that they are effectively trainable using backpropagation. Training of spiking networks is often done indirectly in an offline manner by first training a conventional ANN then mapping the weights to the spiking network~\cite{OConnor_etal13,Diehl_etal15,Cao_etal15,Hunsberger_Eliasmith15}. Recently, several approaches based on approximations to backpropagation have been proposed that can allow online training of spiking networks~\cite{Neftci_etal17,Lee_etal16}. These approaches, however, are based on spiking networks with rate coding, which typically require more memory accesses and longer processing time for each training pattern compared to BSNs. An alternative training approach based on exact backpropagation and temporal coding in spiking networks~\cite{Mostafa_16,Mostafa_etal17} has been shown to lead to highly sparse spiking activity during training and inference, and could potentially be more energetically efficient than training BSNs using the approach presented in this paper.  

In summary, the implementation of binary networks using the proposed architecture uses virtually the same computational resources as a spiking network architecture, while offering significant benefits by reducing memory access and by speeding up learning and inference. BSNs have the attractively low computational overhead of spiking networks, while still being efficiently trainable using backpropagation. They do not incur the increased computational and weight lookup overhead of rate-based spiking networks. Through the use of the approximate pipelined backpropagation scheme outlined in this paper, BSNs can be trained using significantly reduced weight lookup overhead while incurring a modest overhead in the neuron complexity.

\end{document}